\pdfoutput=1

\documentclass[10pt,twocolumn,letterpaper]{article}

\usepackage[pagenumbers]{cvpr} 

\usepackage{times}
\usepackage{latexsym}
\usepackage{amsthm}
\usepackage{amsmath}
\usepackage{booktabs}
\usepackage{multirow} 
\usepackage{makecell}
\usepackage{amsfonts}
\usepackage{hyperref}
\usepackage[T1]{fontenc}

\usepackage[utf8]{inputenc}

\usepackage{microtype}

\usepackage{inconsolata}

\usepackage{graphicx}

\definecolor{waymogreen}{HTML}{00E89D}
\definecolor{waymolgreen}{HTML}{99F7D7} 
\definecolor{waymollgreen}{HTML}{CCFAEB} 
\definecolor{waymoblue}{HTML}{0077FF}
\definecolor{waymolblue}{HTML}{99B7FF}  
\definecolor{waymollblue}{HTML}{CCE4FF} 
\definecolor{waymolgray}{HTML}{F0F0F0}  
\definecolor{mylightgray}{gray}{0.6} 
\usepackage{colortbl}       
\usepackage{pifont}         
\usepackage{footmisc} 

\title{\LARGE \bf
AgentThink\raisebox{-0.1\height}{\includegraphics[width=0.04\linewidth]{ 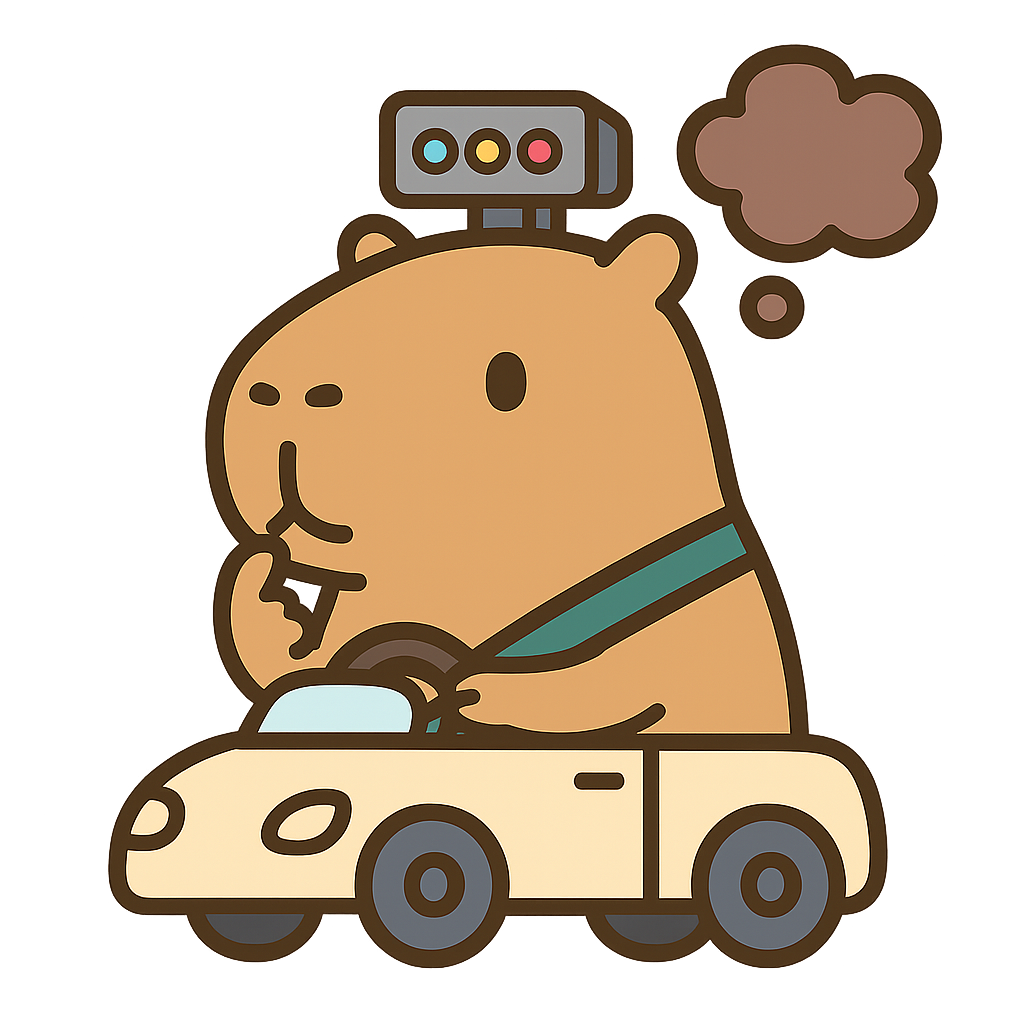}}: A Unified Framework for Tool-Augmented Chain-of-Thought Reasoning in Vision-Language Models for Autonomous Driving}
\author{
 \textbf{Kangan Qian \textsuperscript{1,3,*,\textdaggerdbl}},
 \textbf{Sicong Jiang\textsuperscript{2,*}},
 \textbf{Yang Zhong\textsuperscript{3,*} },
 \textbf{Ziang Luo\textsuperscript{1,3,*}},
 \textbf{Zilin Huang\textsuperscript{4}}, \\
 \textbf{Tianze Zhu\textsuperscript{1}}
 ,
 \textbf{Kun Jiang\textsuperscript{1, \textdagger}},
 \textbf{Mengmeng Yang\textsuperscript{1, \textdagger}},
 \textbf{Zheng Fu\textsuperscript{1}},
 \textbf{Jinyu Miao\textsuperscript{1}},
 \textbf{Yining Shi\textsuperscript{1}},\\
 \textbf{He Zhe Lim\textsuperscript{1}},
 \textbf{Li Liu\textsuperscript{3}},
 \textbf{Tianbao Zhou\textsuperscript{3}}, 
 \textbf{Hongyi Wang\textsuperscript{3}},
 \textbf{Huang Yu\textsuperscript{3}},
 \textbf{Yifei Hu\textsuperscript{3}},\\
 \textbf{Guang Li\textsuperscript{3}},
 \textbf{Guang Chen\textsuperscript{3}},
 \textbf{Hao Ye\textsuperscript{3,\textdagger, \textdaggerdbl}},
 \textbf{Lijun Sun\textsuperscript{2}},
 \textbf{Diange Yang\textsuperscript{1,\textdagger}}
\\
Project website: \href{https://curryqka.github.io/AgentThink.github.io/}{AgentThink Web} \quad Project code: \href{https://github.com/curryqka/AgentThink}{AgentThink Github}
\\
\\
 \textsuperscript{1}School of Vehicle and Mobility, Tsinghua University,
 \textsuperscript{2}McGill University,\\
 \textsuperscript{3}Automotive and Robotics, Xiaomi Corporation,
 \textsuperscript{4}University of Wisconsin-Madison
\\
 \small{
   \textbf{Correspondence:} \href{mailto:qka23@mails.tsinghua.edu.cn}{jiangkun@tsinghua.edu.cn, ydg@tsinghua.edu.cn}
 }
 \thanks{Kangan Qian and Ziang Luo pursued this work during his master's degree at the School of Vehicle and Mobility, Tsinghua University, and finalized it during an internship at Xiaomi.}
 \thanks{Zilin Huang did not receive any funding for this work.}
\thanks{* The authors contribute equally to this work.}
\thanks{\textdagger Corresponding author. \textdaggerdbl Project leader.}
}

\makeatletter
\def\thanks#1{\protected@xdef\@thanks{\@thanks
            \protect\footnotetext{#1}}}
\makeatother
\begin{document}
\maketitle
\begin{abstract}
Vision-Language Models (VLMs) show promise for autonomous driving, yet their struggle with hallucinations, inefficient reasoning, and limited real-world validation hinders accurate perception and robust step-by-step reasoning. To overcome this, we introduce \textbf{AgentThink}, a pioneering unified framework that, for the first time, integrates Chain-of-Thought (CoT) reasoning with dynamic, agent-style tool invocation for autonomous driving tasks. AgentThink's core innovations include: \textbf{(i) Structured Data Generation}, by establishing an autonomous driving tool library to automatically construct structured, self-verified reasoning data explicitly incorporating tool usage for diverse driving scenarios; \textbf{(ii) A Two-stage Training Pipeline}, employing Supervised Fine-Tuning (SFT) with Group Relative Policy Optimization (GRPO) to equip VLMs with the capability for autonomous tool invocation; and \textbf{(iii) Agent-style Tool-Usage Evaluation}, introducing a novel multi-tool assessment protocol to rigorously evaluate the model's tool invocation and utilization. Experiments on the DriveLMM-o1 benchmark demonstrate AgentThink significantly boosts overall reasoning scores by \textbf{53.91\%} and enhances answer accuracy by \textbf{33.54\%}, while markedly improving reasoning quality and consistency. Furthermore, ablation studies and robust zero-shot/few-shot generalization experiments across various benchmarks underscore its powerful capabilities. These findings highlight a promising trajectory for developing trustworthy and tool-aware autonomous driving models. Code is available at \href{https://github.com/curryqka/AgentThink}{\texttt{https://github.com/curryqka/AgentThink}}.

\end{abstract}
\section{Introduction}
\begin{quote}
\textit{"A gentleman is not different by nature, he is good at making use of things around him."}
- Hsun Tzu
\end{quote}

Recent advances in foundation models have opened new opportunities for autonomous driving~\cite{jiang2025survey}, where pretrained Large Language Models (LLMs)~\cite{guo2025deepseek,budzianowski2019hello} and Vision-Language Models (VLMs)~\cite{tian2024drivevlm, zhuang2025math, qian2025fasionad} are increasingly employed to enable high-level scene understanding, commonsense reasoning, and decision making. These models aim to transcend traditional perception pipelines, which rely on hand-crafted components such as object detection~\cite{qian2025lego, streamingflow}, motion prediction~\cite{qian2024priormotion, wang2022sti}, and rule-based planning~\cite{cheng2024pluto} by providing richer semantic representations and broader generalization grounded in web-scale knowledge.

\begin{figure*}[t]
  \centering
  \includegraphics[width=\linewidth]{ 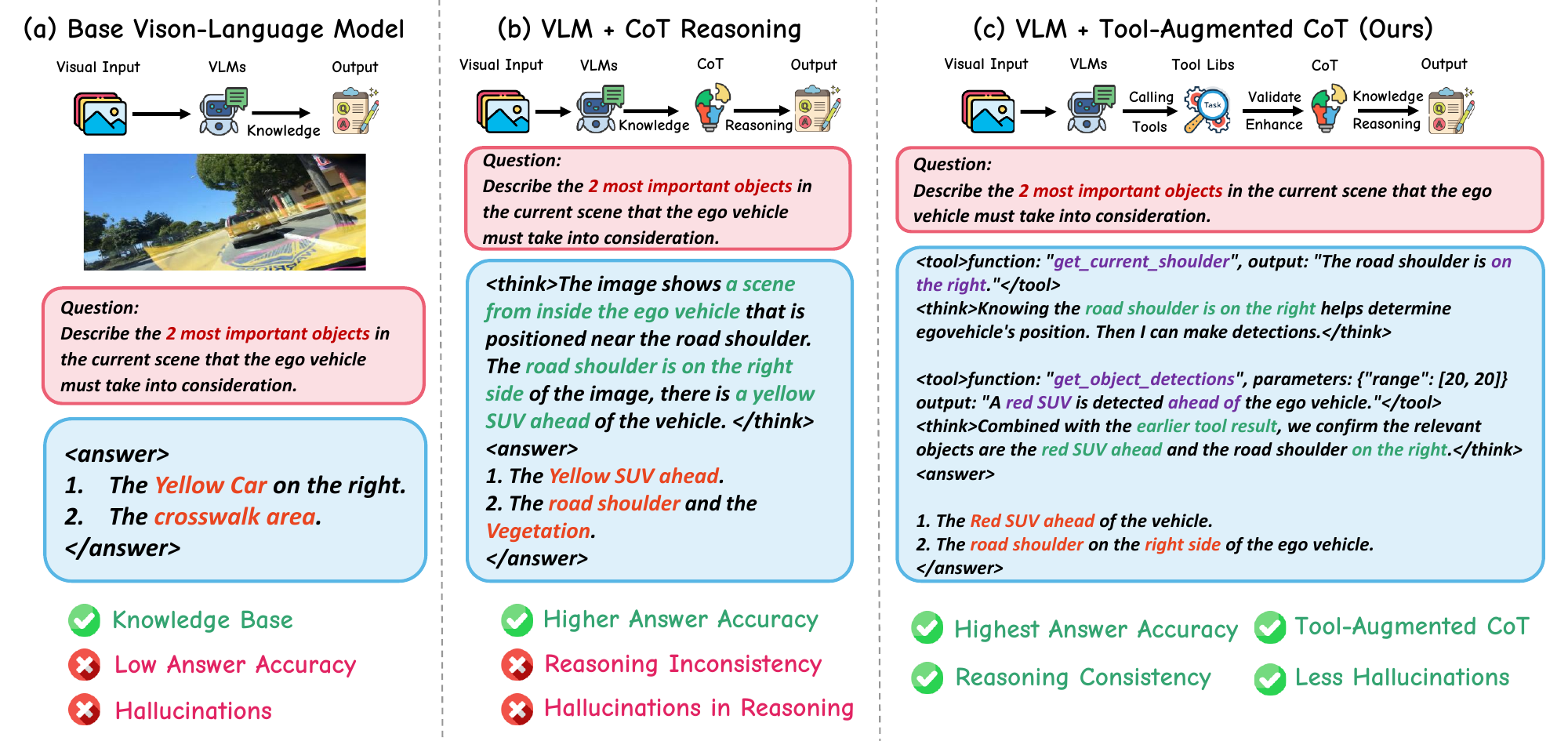}
  \vspace{-5pt}
  \caption{Illustration of the motivation and key highlights of our proposed framework. (a) Base VLMs use static input-output mapping with no reasoning, leading to low accuracy and frequent hallucinations. (b) VLM + CoT introduces structured reasoning, improving interpretability, but still suffers from inconsistencies and lack of verification. (c) AgentThink (Ours) augments CoT with dynamic tool use, enhancing accuracy, reducing hallucinations, and improving reasoning consistency through external verification.}
  \label{fig:motivation}
  \vspace{-10pt}
\end{figure*}

Many recent approaches recast autonomous driving tasks as visual question answering (VQA) problems, applying supervised fine-tuning (SFT) to foundation VLMs with task-specific prompts for object identification, risk prediction, or motion planning~\cite{sima2024drivelm, marcu2024lingoqa, xu2024drivegpt4, ding2024holistic, wang2024omnidrive}. This paradigm are widely used in offline pipelines to mine rare driving scenarios, such as identifying intersections with a typical traffic signals or dense pedestrian activities. However, as shown in Fig.\ref{fig:motivation} (a), these models typically treat reasoning as a static input-to-output mapping, neglecting the uncertainty, and verifiability essential to real-world decision making. Consequently, they often suffer from poor generalization, hallucinated outputs, and limited interpretability~\cite{xie2025vlms}.
\begin{figure}[ht]
  \centering
  \includegraphics[width=\linewidth]{ 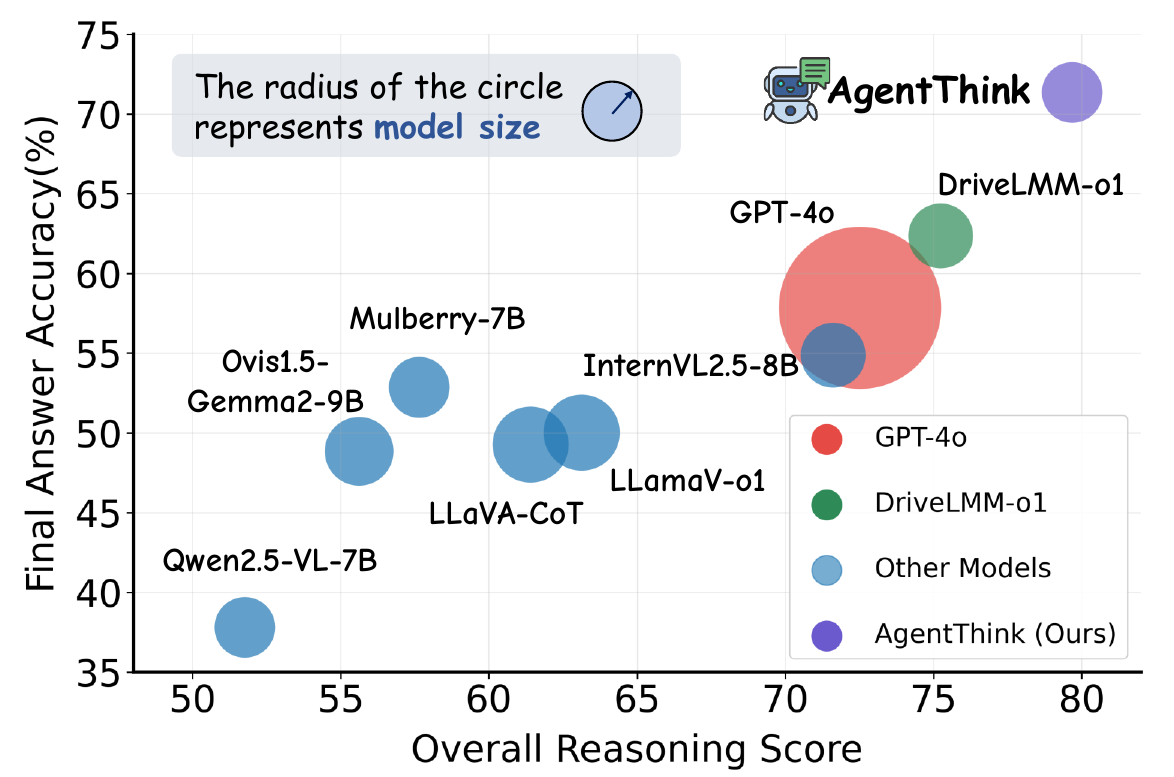}
  \vspace{-5pt}
  \caption{The performance of proposed AgentThink framework on the DriveLMM-o1 benchmark.}
  \label{fig:motivation-result}
  \vspace{-10pt}
\end{figure}

To improve robustness and transparency, recent works have explored incorporating chain-of-thought (CoT) reasoning into VLMs, as shown in Fig. \ref{fig:motivation} (b). Some approaches adopt rigid CoT templates~\cite{tian2024drivevlm, hwang2024emma}, promoting structured logic at the expense of flexibility. Others use open-ended reasoning formats~\cite{nie2024reason2drive, ishaq2025drivelmm}, but may overfit to token patterns and exhibit shallow or redundant reasoning. Moreover, most existing methods rely purely on imitation learning from curated trajectories, lacking the ability to detect knowledge uncertainty or invoke tools for intermediate verification~\cite{zhang2025r1, qian2025fasionad, qian2025fasionad++}.

These challenges lead to a pivotal question: \textit{How can a VLM truly function as a decision-making agent—cognizant of its knowledge boundaries, proficient in verification, and capable of learning from tool-guided feedback?} Inspiration comes from experienced human drivers who will consult aids like mirrors or GPS to refine their judgment when uncertain. Similarly, a capable autonomous agent must not only reason explicitly but also recognize its limitations and dynamically employ tools, such as object detectors or motion predictors, to steer its reasoning and decision-making.

Therefore, we present \textbf{AgentThink}, a unified framework for VLMs in autonomous driving that models reasoning not as a static output, but as an \textit{agent-style process}—in which the model learns to utilize tools to generate Tool-Augmented reasoning chains, verify intermediate steps, and refine its conclusions. As illustrated in Fig.\ref{fig:motivation} (c), rather than blindly mapping inputs to outputs, AgentThink dynamically decides when and how to use tools during inference to support or revise reasoning paths. To enable this behavior, we create a data-training-evaluation pipeline. First, we construct a structured dataset of Tool-Augmented reasoning traces. Then, we introduce a two-stage training pipeline: (i) SFT to warm up reasoning capabilities, and (ii) GRPO \cite{shao2024deepseekmath}, a reinforcement learning (RL)-based strategy that refines reasoning depth and tool-use behavior through structured rewards. Finally, we propose a comprehensive evaluation protocol beyond answer correctness to assess tool selection, integration quality, and reasoning-tool alignment.

As shown in Fig.~\ref{fig:motivation-result}, experiments on the advanced DriveLMM-o1 benchmark~\cite{ishaq2025drivelmm} demonstrate that \textbf{AgentThink} achieves new state-of-the-art performance in both answer accuracy and reasoning score, surpassing existing models. The effectiveness of our approach in cultivating dynamic, tool-aware reasoning is further substantiated by comprehensive ablation studies  and robust generalization capabilities across multiple benchmarks. These collective results strongly suggest that empowering vision-language agents with learned, dynamically invoked tool use is pivotal for creating more robust, interpretable, and generalizable systems for autonomous driving.

\vspace{0.5em}
Generally, our contributions are as follows:
\begin{itemize}
    \item Propose \textbf{AgentThink}, the first framework to integrate dynamic, \textbf{agent-style tool invocation} into vision-language reasoning for autonomous driving tasks. It significantly reduces hallucination in VQA and improves overall reasoning consistency by grounding each reasoning step in reliable tool outputs.

    \item Develop a scalable data generation pipeline that produces \textbf{structured, self-verified} data with \textbf{integrated tool usage} and reasoning chains.

    \item Introduce a two-stage training pipeline that combines \textbf{SFT} with \textbf{GRPO}, enabling models to learn when and how to invoke tools to enhance reasoning performance.

    \item Design new evaluation metrics tailored to \textbf{autonomous driving tool invocation}, capturing tool selection, integration quality, and reasoning-tool alignment. 
\end{itemize}

\section{Related Works}

\subsection{Language Models in Autonomous Driving}

Recent advancements in language modeling have opened up new opportunities for autonomous driving, particularly in enabling interpretable reasoning, commonsense understanding, and decision-making \cite{cui2024survey}. Early efforts integrated LLMs such as GPT series \cite{openai2023gpt4} by recasting driving tasks—e.g., scene description \cite{xu2024drivegpt4, mao2023gpt}, decision-making \cite{fu2024drive,wen2023dilu}, and risk prediction \cite{chen2024driving, ma2024dolphins}—into textual prompts, allowing for zero-shot or few-shot inference. While these approaches showcased the reasoning potential of LLMs, they often lacked step-by-step interpretability and struggled with generalization in out-of-distribution scenarios \cite{wang2023drivemlm}.

Recent works have augmented LLMs with prompting strategies, memory-based context construction, or vision inputs \cite{huang2024vlm}. For instance, DriveVLM \cite{tian2024drivevlm, qian2025fasionad} introduces a CoT approach and dual system with modules for scene description, analysis, and hierarchical planning, while DriveLM \cite{sima2024drivelm} focuses on graph-structured visual question answering. EMMA \cite{hwang2410emma} demonstrates how multimodal models can directly map raw camera inputs to driving outputs, including trajectories and perception objects. Despite these advancements, both LLM-centric and VLM-based methods often treat reasoning as a static input-output mapping, with limited ability to detect uncertainty, perform intermediate verification, or incorporate physical constraints 
 \cite{ishaq2025drivelmm}. Challenges such as hallucinations, over-reliance on rigid templates, and a lack of domain-specific reward feedback persist. To address these limitations, our work introduces a Tool-Augmented, RL-based reasoning framework that enables dynamic and verifiable decision-making for autonomous driving.

\begin{figure*}[t]
  \centering
  \includegraphics[width=\linewidth]{ 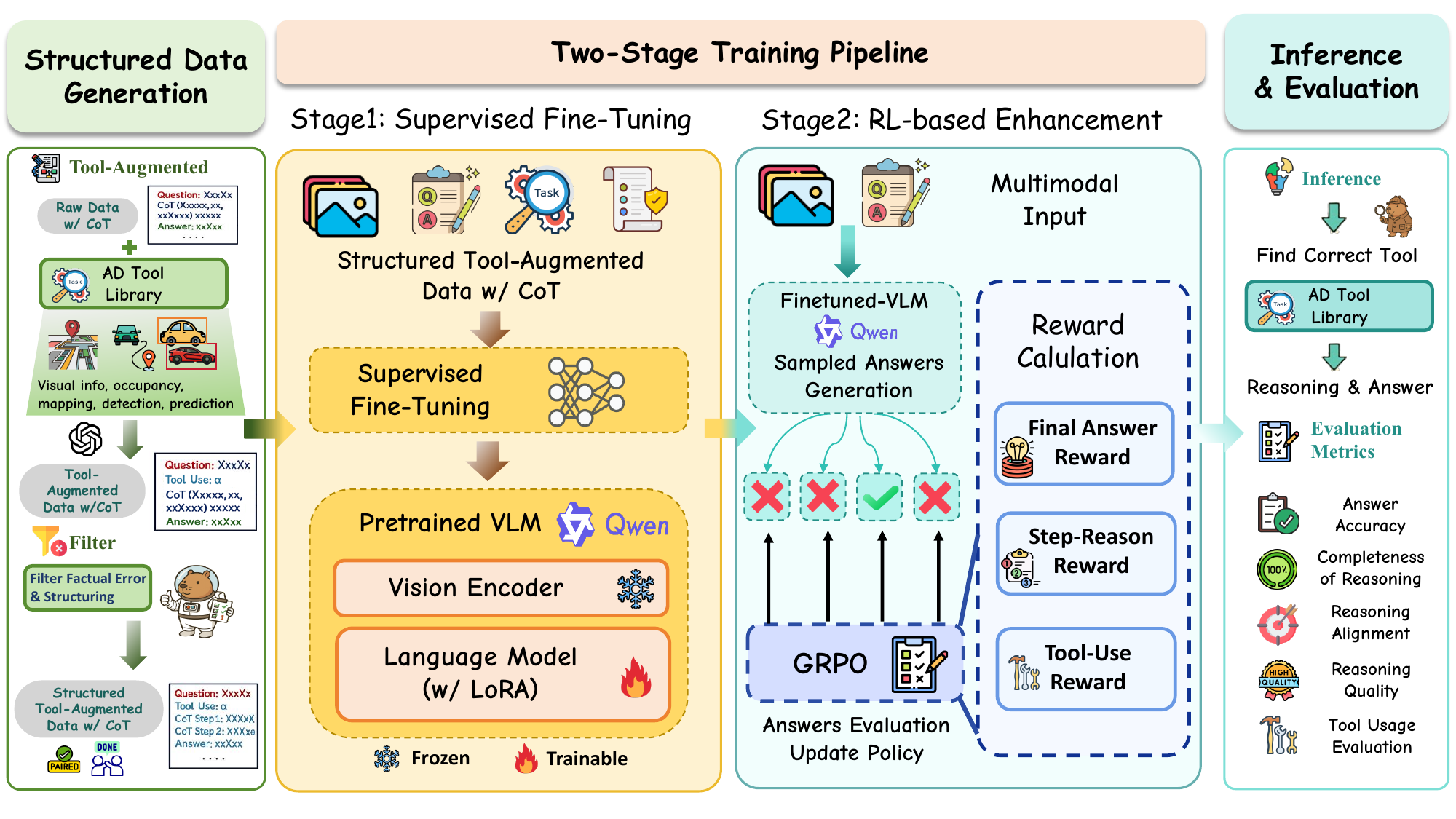}
  \vspace{-20pt}
  \caption{AgentThink Framework Architecture. (i) Structured and scalable data generation pipeline that constructs tool-augmented reasoning; (ii) Two-stage training pipeline that first performs SFT and then applies GRPO  to improve reasoning and tool-use behavior; and (iii) Unified inference and evaluation protocol that dynamically invokes tools and assesses final answers based on reasoning completeness, consistency, and tool-use effectiveness.}
  \label{fig:framework}
  \vspace{-10pt}
\end{figure*}

\subsection{Visual Question Answering in Autonomous Driving}

Visual Question Answering (VQA) for autonomous driving has emerged as a benchmark paradigm to evaluate perception, prediction, and planning capabilities. Benchmarks such as BDD-X \cite{kim2018textual}, DriveBench \cite{xie2025vlms}, DriveMLLM \cite{guo2024drivemllm}, Nuscenes-QA \cite{qian2024nuscenes},  and DriveLMM-o1 \cite{ishaq2025drivelmm} provide structured QA tasks covering complex reasoning scenarios in urban and highway environments. For VQA tasks, recent approaches such as Reason2Drive \cite{nie2024reason2drive},  Alphadrive \cite{jiang2025alphadrive}, OmniDrive \cite{wang2025omnidrive}, and DriveCoT \cite{wang2024drivecot} introduce COT reasoning to enhance model interpretability. 

However, many adopt rigid reasoning templates or rely solely on imitation learning, making them prone to overfitting and hallucination. These methods often overlook dynamic reasoning processes and fail to verify intermediate steps using external tools. In contrast, our framework combines structured data generation with step-level rewards and tool-verification during inference. By employing RL via GRPO, we optimize the model's reasoning trajectory to align with correctness, efficiency, and real-world applicability, setting a new direction for VQA in autonomous driving.



\section{Methodology}
Our framework is designed to address three key challenges: 
(1) \textit{How to generate reasoning data that incorporates tool usage?}
(2) \textit{How to equip the VLM with the capability to effectively utilize tools?} and 
(3) \textit{How to enable reasoning-driven tool invocation during inference, and how to evaluate the model's ability to leverage tools for solving driving-related vision-question-answering tasks?}

Fig. \ref{fig:framework} illustrates AgentThink's three key components: (i) a scalable pipeline for generating structured, tool-augmented reasoning data; (ii) a two-stage training pipeline with SFT and GRPO to improve reasoning and tool-use abilities; and (iii) a new evaluation methodology focused on assessing the model's effective tool utilization and its impact on reasoning.

\subsection{Data Generation Pipeline}\label{sec:data_generation} 
While prior studies \cite{wang2024omnidrive,nie2024reason2drive} have explored reasoning in VLMs, persistent hallucinations remain a challenge. We contend that reliable autonomous driving reasoning, akin to human decision-making, necessitates not only internal knowledge but also the ability to invoke external tools when needed. Addressing this, we introduce a Tool-Augmented data generation pipeline. Unlike existing datasets \cite{wang2023drivemlm,ishaq2025drivelmm} focused solely on reasoning steps and final answers, our pipeline uniquely integrates explicit tool usage into the reasoning process.

\textbf{Tool library.} We developed a specialized tool library inspired by Agent-Driver~\cite{mao2023language}, featuring core modules for  five driving‑centric modules—\textit{visual info}, \textit{detection}, \textit{prediction}, \textit{occupancy}, and \textit{mapping}, plus single‑view vision utilities (open‑vocabulary detection, depth, cropping, resizing). This is augmented by fundamental single-view vision tools like open-vocabulary object detectors and depth estimators. Together, these enable comprehensive environmental information extraction for diverse perception and prediction tasks. Details can be found in the Appendix \ref{sec: Tool}.

\textbf{Prompt Design.} Initial tool-integrated reasoning steps and answers are automatically generated using GPT-4o, guided by a prompt template (as shown in Fig.~\ref{fig:framework}) designed to elicit a tool-augmented reasoning chain for instruction~$\mathcal{L}$ rather than a direct answer.

Specifically, for a pretrained VLM $\pi_{\theta}$, an input image $\mathcal{V}$, and a task instruction $\mathcal{L}$, the reasoning step at time $t$ is generated as:
\begin{equation}
    R_t = \pi_{\theta}(\mathcal{V}, \mathcal{L}, [R_1, \dots, R_{t-1}])
\end{equation}
where $R_t$ denotes the $t$-th reasoning step, and $[R_1, \dots, R_{t-1}]$ represents the previously generated steps in the trajectory. The complete reasoning trajectory is denoted as $\mathcal{T}_R = (R_1, \dots, R_M)$, where $M$ is the maximum number of reasoning steps.

Each reasoning step $R_t$ includes five key elements: the chosen tool ($Tool_i$), a generated sub-question ($Sub_i$), an uncertainty flag ($UF_i$), a guessed answer ($A_i$), and the next action choice ($AC_i$) such as \textit{continue reasoning} or \textit{conclude}. If internal knowledge suffices for $Sub_i$, $A_i$ is outputted and $UF_i = \texttt{False}$; otherwise, $UF_i = \texttt{True}$ and $A_i$ is left blank. This process is repeated to sample $N$ structured reasoning trajectories per QA pair.

\textbf{Data Assessment.} A separate LLM audits each data for factual accuracy and logical consistency, pruning samples with mismatched steps or unsupported conclusions.  The result is a high‑quality corpus that couples explicit tool use with coherent, verifiable reasoning.

\subsection{Two-stage Training Pipeline}\label{sec:training}
After constructing the structured dataset, we design a two-stage training pipeline to progressively enhance the model’s reasoning capabilities and proficiency in tool usage.


\subsubsection{SFT-based Reasoning Warm-up} 
In the first phase, we perform SFT on the Tool-Augmented CoT dataset to warm up the model’s ability to generate structured reasoning chains and appropriate tool calls. Each training sample is represented as $\tau=(\mathcal{V}, \mathcal{L}, \mathcal{T}_R, \mathcal{A})$, where $\mathcal{V}$ is the visual input, $\mathcal{L}$ is the language instruction, $\mathcal{T}_R$ is the step-by-step reasoning trace including both reasoning steps and explicit tool invocation (e.g., \texttt{Tool(name, params)}), and $\mathcal{A}$ is the final answer. The objective is to maximize the likelihood of generating correct reasoning and action sequences:
\begin{equation}
    \mathcal{L}_{\mathrm{SFT}}^{(1)} = -\mathbf{E}_{\tau \sim \mathcal{D}} \sum_{t=1}^{T} \log \pi_{\theta}(R_t \mid \mathcal{V}, \mathcal{L}, R_{<t}),
\end{equation}
where only the generation of reasoning steps and tool calls (action type and parameters) is optimized, while outputs from environment responses (e.g., tool returns) are masked out during loss computation. This phase teaches the model \textit{what tools to use} and \textit{how to configure them} based on the driving context.

In the second phase, we refine the model with full context exposure, including actual tool outputs in the reasoning chain. Training samples now include observed environment feedback $\mathcal{O}$ (e.g., API results, detected objects, or retrieved text), forming an enhanced tuple $(\mathcal{V}, \mathcal{L}, \mathcal{T}_R \cup \mathcal{O}, \mathcal{A})$. The loss remains maximum likelihood but now covers both action generation and observation grounding:
\begin{equation}
    \mathcal{L}_{\mathrm{SFT}}^{(2)} = -\mathbf{E}_{\tau \sim \mathcal{D}} \sum_{t=1}^{T'} \log \pi_{\theta}(z_t \mid \mathcal{V}, \mathcal{L}, z_{<t}),
\end{equation}
where $z_t$ denotes tokens in the extended sequence containing reasoning, actions, and observed outcomes. This joint modeling of actions and observations enables the model to learn an implicit understanding of expected tool outputs, thereby grounding subsequent reasoning in real-world feedback. This two-phase warm-up prepares the model for robust reasoning and effective tool integration prior to RL fine-tuning.

\subsubsection{RLFT-based Reasoning Enhancement}
To further optimize the model beyond imitation learning, we adopt Reinforcement Learning Fine-Tuning (RLFT) using GRPO, which effectively leverages structured rewards without relying on a learned critic.

\textbf{GRPO Overview.}
GRPO avoids the need for a value function by computing the relative advantage of each sample within a group. Given a question $q$ and $G$ responses $\{o_i\}_{i=1}^{G}$ sampled from the old policy $\pi_{\theta_{\text{old}}}$, the GRPO objective is \cite{shao2024deepseekmath}:

{\small
\begin{equation}
\begin{aligned}
\mathcal{J}_{\text{GRPO}}(\theta) = \mathbb{E}_{q,\{o_i\} \sim \pi_{\text{old}}} \Bigg[ 
\frac{1}{G} \sum_{i=1}^G \mathcal{L}_i 
- \beta \, \mathbb{D}_{\text{KL}}(\pi_\theta \| \pi_{\text{ref}}) 
\Bigg]
\end{aligned}
\end{equation}}
where the group-wise clipped loss is defined as:
\begin{equation}
\mathcal{L}_i = \min \left( w_i A_i,\ \operatorname{clip}(w_i,\ 1 - \epsilon,\ 1 + \epsilon) A_i \right)
\end{equation}

and the importance weight $w_i$ and normalized advantage $A_i$ are given by:
\begin{align}
w_i &= \frac{\pi_\theta(o_i \mid q)}{\pi_{\theta_{\text{old}}}(o_i \mid q)} \\
A_i &= \frac{r_i - \operatorname{mean}(r)}{\operatorname{std}(r)}
\end{align}
where $r_i$ denotes the reward assigned to output $o_i$, and $\beta$ and $\epsilon$ are tunable hyperparameters.

\textbf{Reward Design.}
To guide the model toward accurate, interpretable, and tool-aware reasoning, we design a structured reward function with three major components: 


\begin{table}[h]
\centering
\small
\label{tab:tool_rewards}
\begin{tabular}{p{0.24\linewidth} p{0.62\linewidth}}
\toprule
\textbf{Reward Type} & \textbf{Description} \\
\midrule
Final Answer Reward & Verifies final answer against ground truth; promotes task-level correctness. \\
\midrule
Step Reasoning Reward & Evaluates intermediate step logic \& structure. Sub-rewards for: \newline
(i) Step Matching: Align with reference steps and penalize incorrect ordering.  \newline
(ii) Coherence: Smooth, logical transitions between steps. \\
\midrule
Tool Use Reward & Promotes appropriate \& meaningful tool usage. Sub-rewards for: \newline
(i) Format Compliance: Adherence to expected output structure (e.g., "Tool", "Step Reasoning"). \newline
(ii) Integration Quality: Effective coherent incorporation of tool outputs into reasoning. \\
\bottomrule
\end{tabular}
\vspace{-5pt}
\caption{GRPO reward for tool-augmented reasoning}
\vspace{-10pt}
\end{table}

This structured reward design provides more targeted and interpretable supervision than generic similarity metrics. It enables GRPO to optimize both the quality of the reasoning process and the model’s ability to invoke tools when needed.

\begin{figure}[t]
  \centering
  \includegraphics[width=\linewidth]{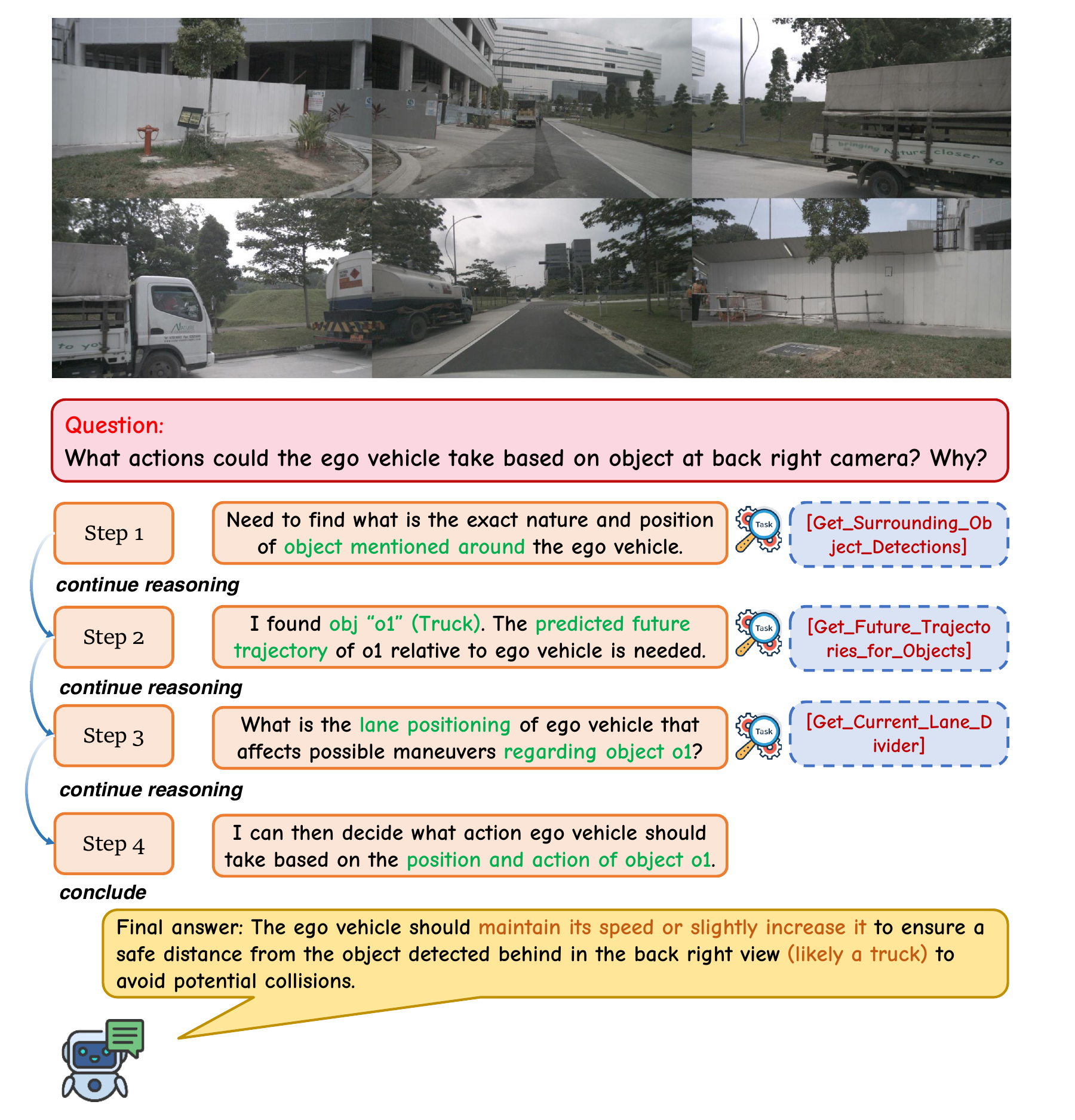}
  \vspace{-10pt}
  \caption{The model generates structured reasoning chains, dynamically invokes external tools to resolve uncertainties (e.g., object detection, trajectory prediction, lane width), and concludes with an interpretable action recommendation.}
  \label{fig:infer}
  \vspace{-10pt}
\end{figure}



\subsection{Inference and Evaluation}\label{sec:inference}
During inference, as illustrated in Fig.~\ref{fig:infer}, the VLM dynamically accesses tools from a predefined library to acquire relevant information, enabling step-by-step reasoning. This dynamic tool-calling mechanism not only improves answer accuracy but also mirrors the structure of our tool-augmented training data. However, existing benchmarks \cite{guo2024drivemllm, ishaq2025drivelmm} do not evaluate tool-usage capabilities. To bridge this gap, we introduce three metrics (Table~\ref{tab:tool_metrics}) that assess the model’s tool utilization during reasoning, leveraging the LLM-as-Judge paradigm \cite{jiang2025survey}.

\begin{table}[ht]
\centering
\small
\begin{tabular}{p{0.24\linewidth} p{0.6\linewidth}}
\toprule
\textbf{Metric} & \textbf{Description} \\
\midrule
Tool Usage Appropriateness & Assesses whether tools are logically selected and meaningfully used to support individual reasoning steps. \\
\midrule 
Tool Chain Coherence & Evaluates whether the sequence of tool invocations is clear, logically ordered, and efficiently contributes to reasoning. \\
\midrule 
Perception-Guided Tool Alignment & Measures how well tool usage aligns with multimodal inputs, including visual observations and scene context. \\
\bottomrule
\end{tabular}
\caption{Evaluation metrics for tool-use in reasoning}
\label{tab:tool_metrics}
\vspace{-10pt}
\end{table}

\section{Experiments}

\begin{table*}[!t]
\centering
\setlength{\tabcolsep}{3pt}
\small
\resizebox{\linewidth}{!}{
\begin{tabular}{l|ccc|cc|cc}
    \toprule
    \multirow{2}{*}{Vision Language Models} 
        & \multicolumn{3}{c|}{Driving Metrics (\%) $\uparrow$} 
        & \multicolumn{2}{c|}{Scene Detail (\%) $\uparrow$} 
        & \multicolumn{2}{c}{Overall(\%) $\uparrow$} \\ 
        & Risk Assess. & Rule Adh. & Scene Aware. 
        & Relevance & Missing 
        & Reason. & MCQ \\ 
    \midrule
     
    GPT-4o \cite{islam2024gpt} &71.32 &80.72 &72.96 &76.65& 71.43 &72.52 &57.84\\
    Ovis1.5-Gemma2-9B \cite{lu2024ovis}                & 51.34 & 66.36 & 54.74 & 55.72 & 55.74 & 55.62 & 48.85 \\
    Mulberry-7B \cite{yao2024mulberry}                     & 51.89 & 63.66 & 56.68 & 57.27 & 57.45 & 57.65 & 52.86 \\
    LLaVA-CoT \cite{xu2024llava}                       & 57.62 & 69.01 & 60.84 & 62.72 & 60.67 & 61.41 & 49.27 \\
    LlamaV-o1 \cite{thawakar2025llamav}                       & 60.20 & 73.52 & 62.67 & 64.66 & 63.41 & 63.13 & 50.02 \\
    InternVL2.5-8B \cite{chen2024expanding}                & 69.02 & 78.43 & 71.52 & 75.80 & 70.54 & 71.62 & 54.87 \\
    Qwen2.5-VL-7B \cite{bai2025qwen25vltechnicalreport}                   & 46.44 & 60.45 & 51.02 & 50.15 & 52.19 & 51.77 & 37.81 \\
    DriveLMM-o1 \cite{ishaq2025drivelmm}                    & 73.01 & 81.56 & 75.39 & 79.42 & 74.49 & 75.24 & 62.36 \\
    \cellcolor{waymolgray}\textbf{AgentThink (Ours)} & \cellcolor{waymolgray}\textbf{80.51} & \cellcolor{waymolgray}\textbf{84.98} & \cellcolor{waymolgray}\textbf{82.11} & \cellcolor{waymolgray}\textbf{84.99} & \cellcolor{waymolgray}\textbf{79.56} & \cellcolor{waymolgray}\textbf{79.68} & \cellcolor{waymolgray}\textbf{71.35} \\
    \bottomrule
\end{tabular}}
\vspace{-5pt}
\caption{Evaluation results on the DriveLMM-o1 benchmark. AgentThink significantly improves reasoning and answer accuracy across all categories by leveraging dynamic Tool-Augmented reasoning.}
\label{tab:main-table}
\end{table*}

In this section, we conduct extensive experiments to validate the effectiveness of AgentThink. Our experiments are designed to answer the following core questions:

\begin{itemize}
    \item[\textbf{Q1.}] Can dynamic Tool-Augmented reasoning improve both final answer accuracy and reasoning consistency over existing VLM baselines?
    \item[\textbf{Q2.}] Does our structured reward design (Final Answer, Step-wise Reason, Tool-use) contribute meaningfully to reasoning behavior?
    \item[\textbf{Q3.}] How well does AgentThink generalize to unseen datasets under zero-shot and one-shot settings?
\end{itemize}

\begin{table*}[!t]
\centering
\setlength{\tabcolsep}{4pt}
\small
\resizebox{\linewidth}{!}{
\begin{tabular}{l|c|ccc|ccc|cc|cc}
    \toprule
    \multirow{2}{*}{Model Variant}
    & SFT & \multicolumn{3}{c|}{GRPO Reward Setting} 
    & \multicolumn{3}{c|}{Driving Metrics (\%) $\uparrow$} 
    & \multicolumn{2}{c|}{Scene Detail (\%) $\uparrow$} 
    & \multicolumn{2}{c}{Overall (\%) $\uparrow$} \\
    & Setting & Answer & Step R. & Tool Use
    & Risk Assess. & Rule Adh. & Obj Und.
    & Relevance & Missing 
    & Reason. & MCQ \\
    \midrule
    Base Model & \ding{55}                       & \ding{55} & \ding{55} & \ding{55} & 46.44   & 60.45   & 51.02   & 50.15   & 52.19   & 51.77   & 37.81 \\
    + SFT& \cellcolor{waymollblue}\checkmark   & \ding{55} & \ding{55} & \ding{55} & 70.25 & 79.83     & 75.41      & 81.45   & 71.68   & 72.54   & 62.95  \\
    + GRPO & \ding{55} & \cellcolor{waymollblue}\checkmark & \ding{55} & \ding{55} & 69.25   & 75.41   & 71.58   & 75.86   & 68.05   & 69.41   & 61.41 \\
    + GRPO\textdagger & \ding{55} & \ding{55} & \cellcolor{waymollblue}\checkmark & \ding{55} & 69.29   & 75.43   & 72.66   & 76.77   & 69.03   & 69.43   & 57.19 \\
    + SFT + GRPO   &  \cellcolor{waymollblue}\checkmark & \cellcolor{waymollblue}\checkmark & \cellcolor{waymollblue}\checkmark & \ding{55} & 71.00   & 77.35   & 73.23   & 78.13   & 69.08   & 70.83      & 64.58 \\
    \textbf{AgentThink (Ours)}  &  \cellcolor{waymollblue}\checkmark & \cellcolor{waymollblue}\checkmark & \cellcolor{waymollblue}\checkmark & \cellcolor{waymollblue}\checkmark & \textbf{80.51} & \textbf{84.98} & \textbf{82.11} & \textbf{84.99} & \textbf{79.56} & \textbf{79.68} & \textbf{71.35} \\
    \bottomrule
\end{tabular}
}
\vspace{-5pt}
\caption{Ablation study of AgentThink on reward design and training strategy. The full model (bottom row) benefits from the combination of SFT, GRPO, and structured tool-use rewards. Ablation models here trained with 8 epochs.}
\vspace{-10pt}
\label{tab:ablation-table}
\end{table*}
\begin{table}[htbp]
\centering
\setlength{\tabcolsep}{4pt}
\small
\resizebox{\linewidth}{!}{
\begin{tabular}{lcccc}
\toprule
\textbf{Model} & \makecell[c]{Tool Usage \\ Appro.} & \makecell[c]{Tool Chain \\ Coh.} & \makecell[c]{Percep-Guided \\ Tool Align.} & \makecell[c]{Overall Tool \\ Score} \\
\midrule
Base + DirectTool & 59.61 & 73.29 & 69.71 & 67.54 \\
Base + SFT & 62.38 & 78.19 & 75.78 & 72.12 \\
Base + GRPO & 68.44 & 80.73 & 80.82 & 76.66 \\
\cellcolor{waymolgray}\textbf{AgentThink (Ours)} & \cellcolor{waymolgray}\textbf{70.92} & \cellcolor{waymolgray}\textbf{82.16} &\cellcolor{waymolgray}\textbf{ 84.25}&\cellcolor{waymolgray}\textbf{ 79.11} \\
\bottomrule
\end{tabular}}
\vspace{-5pt}
\caption{Tool Evaluation Results of Different Qwen2.5-VL-7B Variants.}
\vspace{-10pt}
\label{tab:tool-use}
\end{table}
\textbf{Evaluation Metrics.}
We leverage DriveLMM-o1's evaluation metrics, specifically utilizing the overall reasoning score to gauge the reasoning of VLMs, and employing multiple choice quality (MCQ) to assess the accuracy of the final answers, with further details provided in the Appendix \ref{sec:Evaluation Metric}. Additionally, we introduce new metrics to evaluate tool-use capability as described in Table~\ref{tab:tool_metrics}.

\textbf{Model and Implementation.} We employ Qwen2.5-VL-7B as our base model and keep its vision encoder frozen. Supervised fine-tuning (SFT) is applied via LoRA for 20 epochs with a learning rate of $1 \times 10^{-4}$, followed by GRPO fine-tuning for 8 epochs with a learning rate of $1 \times 10^{-5}$. The training batch size is set to 1 per device. We use the \texttt{bfloat16} data type to improve computational efficiency. All experiments are conducted on 16$\times$ NVIDIA A800 GPUs. During the GRPO phase, we perform 2 rollouts per question. Additional implementation details are provided in Appendix~\ref{sec:app-implement}.
\subsection{Main Experiment Results}

\textbf{Comparison with Open-Source VLMs.}
Table~\ref{tab:main-table} presents the main results on the DriveLMM-o1 benchmark, comparing AgentThink with a range of strong open-source VLM models, including DriveLMM-o1\cite{ishaq2025drivelmm}, InternVL2.5~\cite{chen2024expanding}, LLaVA-CoT \cite{xu2024llava}, and Qwen2.5-VL variants.

Our full model, AgentThink, achieves state-of-the-art performance across all categories. It surpasses the baseline Qwen2.5-VL-7B by a wide margin, improving the overall reasoning score from 51.77 to 79.68 (+53.91\%), and final answer accuracy from 37.81\% to 71.35\% (+33.54\%). Compared to the strongest prior system, DriveLMM-o1, which already integrates some reasoning abilities, AgentThink further improves by +5.9\% in reasoning and +9.0\% in final answer accuracy—demonstrating the advantage of learned tool-use over static CoT or imitation-based methods.

\textbf{Performance Breakdown.}
In addition to reasoning and accuracy, AgentThink consistently outperforms others in driving-specific metrics (risk assessment, traffic rule adherence, and scene understanding), as well as perception-related categories (relevance and missing detail detection). These gains reflect its ability to leverage dynamic tool invocation and feedback to ground its reasoning more effectively in visual context.

\textbf{Key Insight.}
Unlike traditional CoT or prompt-based methods, AgentThink learns \textit{when} and \textit{why} to invoke external tools, enabling more adaptive and context-aware reasoning. This leads to better decision quality, fewer hallucinations, and improved trustworthiness in safety-critical driving scenarios. We provide the case in the Appendix \ref{sec:Visualization}.
\begin{figure*}[t]
  \centering
  \includegraphics[width=\linewidth]{ 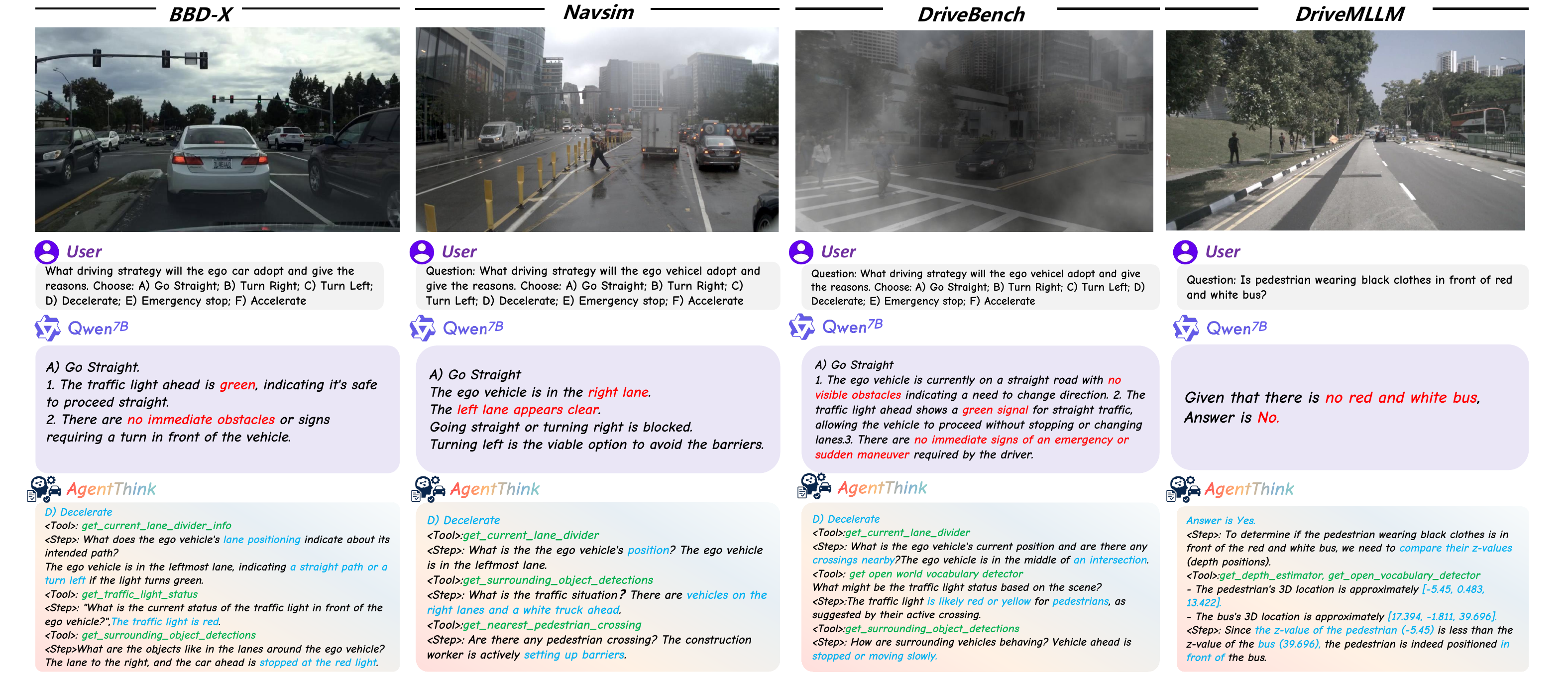}
  \caption{Zero-shot qualitative comparison with Qwen2.5VL-7B on BDD-X, Navsim, DriveBench and DriveMLLM.}
  \label{fig:case2}
\end{figure*}

\begin{table*}[!t]
\centering
\setlength{\tabcolsep}{3pt}
\small
\begin{tabular}{l|l|cccc|cccc|c|c}
    \toprule
    \multirow{2}{*}{Type} & \multirow{2}{*}{Model} 
        & \multicolumn{9}{c|}{Metric Scores (\%) $\uparrow$} 
        & \multirow{2}{*}{\textbf{Overall}} \\
        \cmidrule(lr){3-11}
        & & L/R & F/B & RHD & RD & PPos & BBox & CVD & CD & AccS \\ 
    \midrule
    \multirow{7}{*}{Zero-shot} 
        &GPT-4o \cite{islam2024gpt} & 91.72 & 67.60 & 9.58 & 14.69 & 40.90 & 4.07 & 46.11 & 70.65 & 43.16 & 25.63 \\ 
        &GPT-4o-mini & 67.67 & 50.13 & 70.44 & 0.00 & 29.28 & 3.78 & 0.00 & 46.40 & 33.46 & 16.68 \\ 
        & LLaVA-ov-72B \cite{li2024llava} & 85.42 & 49.48 & 13.76 & 45.27 & 16.46 & 0.00 & 42.97 & 27.09 & 35.06 & 21.10 \\
        & Qwen2.5-VL-7B \cite{bai2025qwen25vltechnicalreport}            & 76.55 & 55.24 &  7.14 & 17.11 & 55.97 & 38.31 & 55.94 & 51.52 & 44.72 & 13.36 \\
        & Qwen + CoT     & \textbf{87.06} & \textbf{63.09} & 16.69 & 22.56 & 52.51 & 38.87 & \textbf{76.90} & 38.71 & 49.55 & 19.31 \\
        & Qwen + DirectTool   & 78.95 & 48.96 & \textbf{58.43} & \textbf{67.57} & 58.20 & 42.22 & 51.76 & 51.38 & \textbf{57.18} & 24.05 \\
        & \cellcolor{waymolgray}\textbf{AgentThink (Ours)} & \cellcolor{waymolgray}82.33 & \cellcolor{waymolgray}54.40 & \cellcolor{waymolgray}56.14 & \cellcolor{waymolgray}61.45 & \cellcolor{waymolgray}\textbf{70.45} & \cellcolor{waymolgray}\textbf{56.23} & \cellcolor{waymolgray}23.09 & \cellcolor{waymolgray}\textbf{51.60} & \cellcolor{waymolgray}56.96 & \cellcolor{waymolgray}\textbf{26.52} \\ 
    \midrule
    \multirow{7}{*}{One-shot} 
        & GPT-4o & 91.08 & 69.37 & 36.51 & 71.17 & 42.44 & 5.10 & 0.00 & 63.88 & 47.44 & 33.17 \\ 
        & GPT-4o-mini & 66.00 & 48.95 & 83.02 & 58.47 & 25.71 & 3.97 & 52.73 & 55.23 & 49.26 & 22.13 \\ 
        & LLaVA-ov-72B\cite{li2024llava} & 79.12 & 62.97 & 49.26 & 68.04 & 28.57 & 2.20 & 53.12 & 60.90 & 50.52 & 36.66 \\
        & Qwen2.5-VL-7B \cite{bai2025qwen25vltechnicalreport}   & 80.30 & 53.14 & 36.96 & 39.13 & 62.69 & 22.63 & 49.88 & 48.32 & 49.13 & 33.53 \\
        & Qwen + CoT    & \textbf{86.35} & \textbf{59.95} & 43.29 & 31.81 & 53.64 & 26.93 & 51.02 & 42.30 & 49.41 & 32.06 \\
        & Qwen + DirectTool   & 84.57 & 55.50 & \textbf{67.32} & 59.54 & \textbf{85.58} & 26.07 & \textbf{52.34} & \textbf{53.25} & 60.52 & 42.27 \\
        & \cellcolor{waymolgray}\textbf{AgentThink (Ours)} & \cellcolor{waymolgray}78.71 & \cellcolor{waymolgray}48.46 & \cellcolor{waymolgray}60.64 & \cellcolor{waymolgray}\textbf{60.71} & \cellcolor{waymolgray}72.36 & \cellcolor{waymolgray}\textbf{64.46} & \cellcolor{waymolgray}52.26 & \cellcolor{waymolgray}52.04 & \cellcolor{waymolgray}\textbf{61.21} & \cellcolor{waymolgray}\textbf{47.24} \\
    \bottomrule
\end{tabular}
\vspace{-5pt}
\caption{Zero-shot and one-shot performance comparison across multiple metrics on the DriveMLLM benchmark.}
\vspace{-10pt}
\label{tab:shot_comparison}
\end{table*}

\subsection{Tool-Use Analysis}

As mentioned above, we analyze how different training strategies influence tool-use behavior during reasoning. Table~\ref{tab:tool-use} reports results on these three dimensions: (1) Tool usage appropriateness, (2) Tool chain coherence, and (3) Perception-guided alignment. 

The DirectTool baseline, which enforces tool invocation via prompt without reasoning structure, shows moderate chain coherence but lower appropriateness and alignment—indicating that forced tool use often lacks purpose. Adding SFT improves both appropriateness and alignment, but lacks feedback on tool quality, limiting further gains. GRPO with structured rewards leads to significant improvements, teaching the model to invoke tools selectively and integrate outputs coherently. Our full model, combining SFT and GRPO with full reward, achieves the best performance across all metrics. This demonstrates that both supervision and reward shaping are essential for learning effective, context-aware tool use. We also evaluate the impact of the training data scale, as detailed in Appendix \ref{sec: Impact}.



\begin{table*}[!ht]
\centering
\setlength{\tabcolsep}{5pt}
\small
\begin{tabular}{l|l|l|ccccc}
\toprule
\multirow{2}{*}{Model} & \multirow{2}{*}{Size} & \multirow{2}{*}{Type}
  & \multicolumn{5}{c}{BenchDrive Scores (\%) $\uparrow$} \\
\cmidrule(lr){4-8}
 & & & Weather & External & Sensor & Motion & Transmission \\
\midrule
GPT\text-4o \cite{islam2024gpt}              & \textemdash & Commercial & 57.2 & 29.3 & 44.3 & 34.3 & 36.8 \\
LLaVA\text-1.5 \cite{li2024llava}          & 7 B         & Open       & 69.7 & 26.5 & 18.8 & 71.3 & 10.2 \\
LLaVA\text-1.5            & 13 B        & Open       & 61.6 & 15.5 & 24.1 & \textbf{79.8} & 15.5 \\
LLaVA\text-NeXT           & 7 B         & Open       & 69.7 & 48.5 & 21.8 & 66.0 & 11.8 \\
InternVL2    \cite{chen2024expanding}             & 8 B         & Open       & 59.9 & 50.8 & 29.9 & 68.3 & 30.0 \\
Phi\text-3                & 4.2 B       & Open       & 40.0 & 25.0 & 16.8 & 31.3 & 27.7 \\
Phi\text-3.5              & 4.2 B       & Open       & 60.6 & 21.3 & 25.6 & 33.0 & 39.7 \\
Qwen2\text-VL \cite{bai2025qwen25vltechnicalreport}           & 7 B         & Open       & \textbf{76.7} & 37.5 & 22.8 & 57.0 & 35.8 \\
Qwen2\text-VL             & 72 B        & Open       & 59.8 & 45.5 & 52.3 & 58.3 & 44.8 \\
DriveLM   \cite{sima2024drivelm}                & 7 B         & Specialist & 21.2 & 21.3 &  9.0 & 22.3 & 17.5 \\
Dolphins   \cite{ma2024dolphins}               & 7 B         & Specialist & 54.3 & 30.0 &  9.4 &  9.3 & 21.5 \\
\rowcolor{waymolgray}
\textbf{AgentThink (Ours)} & 7 B       & Ours   & 64.8 & \textbf{68.2} & \textbf{56.8} & 71.8 & \textbf{61.2} \\
\bottomrule
\end{tabular}
\vspace{-4pt}
\caption{DriveBench results. AgentThink achieves SOTA on 3/5 tasks (External, Sensor, Transmission), and ranks top-3 on the remaining 2 (Weather, Motion), indicating strong generalization.}
\label{tab:benchdrive_results}
\end{table*}
\subsection{Ablation Study}

In Table~\ref{tab:ablation-table}, we conduct comprehensive ablations to examine the effect of different reward signals and training stages in AgentThink. Using SFT or GRPO alone with either final-answer or step-reasoning reward provides moderate gains over the base model, improving task accuracy and reasoning coherence respectively. However, their effects are limited when applied in isolation.

We find that combining SFT with GRPO (without the tool-use reward) delivers better performance, which shows that warm-up reasoning is crucial before reinforcement tuning. Our complete AgentThink model, which incorporates all three reward components, attains the optimal results. It greatly enhances both reasoning quality and answer accuracy, thus emphasizing the importance of using tools and grounding reasoning in visual context.

\subsection{Generalization Evaluation}
We assessed AgentThink's generalization ability on a new DriveMLLM benchmark under zero-shot and one-shot settings against a range of strong baselines, including prominent VLMs and task-specific variants (details in Table~\ref{tab:shot_comparison}). The evaluation metrics are detailed in Appendix~\ref{sec: Evaluation DriveMLLM}.

AgentThink achieves state-of-the-art performance with \textbf{zero-shot} (26.52) and \textbf{one-shot} (47.24) scores, surpassing GPT-4o and LLaVA-72B. While baseline methods like \texttt{DirectTool} demonstrate strong perception task results (\textit{e.g.}, \texttt{RHD} 89.2 vs.~86.1, \texttt{BBox} precision 92.4\% vs.~91.7\%) through hard-coded tool prompts, they suffer from contextual rigidity and fragmented reasoning-perception alignment. Our model demonstrates a superior balance by effectively coordinating explicit reasoning with learned, adaptive tool use grounded in perceptual context. This underscores the advantages of its learned tool-use mechanism over static prompting or sheer model scale for robust generalization.

Shown in Table~\ref{tab:benchdrive_results}, in the zero-shot experiments on DriveBench~\cite{xie2025vlms}, AgentThink achieved state-of-the-art performance on 3 out of 5 tasks (External, Sensor, and Transmission) and ranked within the top 3 on the remaining two (Weather and Motion). This demonstrates its strong generalization ability, showing that it can maintain high performance across diverse unseen tasks and scenarios, surpassing both general-purpose and specialist models in Driving VQA.

Qualitatively, as illustrated in Fig.~\ref{fig:case2}, AgentThink successfully navigates challenging zero-shot corner cases on diverse benchmarks (BDD-X~\cite{kim2018textual}, Navsim~\cite{dauner2024navsim}, DriveBench~\cite{xie2025vlms}, DriveMLLM~\cite{guo2024drivemllm}). In these cases, the base Qwen model often fails to gather sufficient information or generates hallucinates during reasoning, leading to incorrect outputs. In contrast, AgentThink adeptly invokes tools to acquire critical decision-making information, thereby correctly answering these difficult questions. This further highlights the practical benefits of its dynamic, tool-augmented reasoning in unfamiliar contexts.

\section{Conclusion}
We present \textbf{AgentThink}, the first unified framework that tightly fuses CoT reasoning with agent‑style tool invocation for autonomous driving. With a scalable tool‑augmented dataset and a two‑stage SFT with GRPO pipeline, AgentThink raises DriveLMM‑o1 reasoning score from \textbf{51.77} to \textbf{79.68} and answer accuracy from \textbf{37.81\%} to \textbf{71.35\%}, outperforming the strongest prior model by +5.9\% and +9.0\%. Beyond improved performance, AgentThink demonstrates stronger interpretability by making each reasoning step grounded in tool outputs. Notably, as a driving-scene VQA system, AgentThink aligns with industrial practices where such models are used off the control loop, for example, for mining corner cases or providing high-level feedback in a dual-system setup, and thus operating under relaxed latency constraints compared to real-time planning. Results validate that coupling explicit reasoning with learned tool use is a promising path toward safer, more robust language-model‑centric driving tasks. We believe this framework lays the foundation for building trustworthy VLM-based agents capable of generalizing to complex, dynamic real-world driving environments.

\section*{Acknowledgement}
\noindent This work was supported in part by the National Natural Science Foundation of China (52372414, 52394264, 52472449).
Zilin Huang did not receive any funding for this work. 
We extend our gratitude to Xiaomi for their contribution to this work. Xiaomi generously provided computational resources that significantly accelerated our experiments and model training.  Additionally, their supportive internship environment enabled close collaboration between academic and industrial researchers, fostering innovation in real-world autonomous driving challenges.

\section*{Limitations}

\textbf{Data scale.} Our tool‑augmented corpus totals 18k annotated instances, limiting exposure to long‑tail or rare driving events. A substantially larger and more diverse dataset is required for the model to internalize a broader spectrum of real‑world scenarios.

\textbf{Model size.} Ours relies on \emph{qwen2.5‑VL‑7B}; the 7B‑parameter footprint incurs non‑trivial memory and latency overhead on embedded automotive hardware. Future work should investigate lighter backbones (e.g., \textasciitilde3B) that preserve reasoning capability while easing on‑board resource constraints.

\textbf{Lack of temporal context.} The model under discussion processes single-frame, multi-view images as inputs. However, in the absence of sequential information, it may misinterpret scenarios that rely on temporal cues, such as changing traffic lights. To address this issue, incorporating video tokens or employing recurrent memory could be effective solutions.

\textbf{Missing 3‑D modalities.} The absence of LiDAR or point‑cloud data deprives the model of precise spatial geometry, introducing uncertainty in distance‑related reasoning. Fusing additional modalities is expected to enhance robustness.

\section*{Ethics Statement}

All data come from publicly released driving datasets that anonymise personally identifiable information; no private or crowd‑sourced data were collected. The study involves no human subjects, and every experiment is run offline or in simulation. Model checkpoints are released under a non‑commercial licence that prohibits deployment in safety‑critical vehicles without additional validation. The work follows the ACL Code of Ethics and does not rely on sensitive data or models.

{
    \small
    \bibliographystyle{ieee_fullname}
    \bibliography{acl_latex}
}

\clearpage

\appendix

\section*{Appendix}
\label{sec:appendix}

\section{Data Generation Pipeline}
\subsection{Tool Library}
\label{sec: Tool}
This tool library is designed to support autonomous driving function calls by providing a comprehensive set of tools for various tasks including visual detection, object detection, trajectory prediction, map information querying, and more. Below is an introduction to the main categories and specific tools available in the library:
\paragraph{Visual info Functions}
\begin{itemize}
    \item \textbf{get\_open\_world\_vocabulary\_detection}: Given a list of object words, this function detects the objects in the image and returns their 2D positions and sizes within the camera coordinate system. If no related object is found, it returns None.
    \item \textbf{get\_3d\_loc\_in\_cam}: Given an input image and object-related keywords, this function calculates the depth value for each pixel and determines the 3D locations of the specified objects within the camera coordinate system.
    \item \textbf{Resize\_image\_info}: defines a function to resize an image to specified dimensions, supporting various interpolation methods. It requires the input image path, output path, and target size with, interpolation method as an optional parameter.
    \item \textbf{Crop\_image\_info}: defines a function to crop a rectangular region from an image. It requires the input image path, output path, and the coordinates of the crop region.
\end{itemize}

\paragraph{Detection Functions}
\begin{itemize}
    \item \textbf{get\_leading\_object\_detection}: Detects the leading object, returning its ID, position, and size. If no leading object exists, it returns None.
    \item \textbf{get\_surrounding\_object\_detections}: Detects surrounding objects within a 20m×20m range, providing a list of object IDs along with their positions and sizes. Returns None if no surrounding objects are present.
    \item \textbf{get\_front\_object\_detections}: Identifies objects within a 10m×20m range in front of the vehicle, returning their IDs, positions, and sizes. Returns None if no such objects exist.
    \item \textbf{get\_object\_detections\_in\_range}: Detects objects within a specified range (x\_{start}, x\_{end})×(y\_{start}, y\_{end})m², returning a list of object IDs and their corresponding positions and sizes. Returns None if no objects are found in the range.
    \item \textbf{get\_all\_object\_detections}: Retrieves detections for all objects in the entire scene, providing a list of object IDs and their positions and sizes. However, it is recommended to avoid using this function if more specific alternatives are available.
\end{itemize}

\paragraph{Prediction Functions}
\begin{itemize}
    \item \textbf{get\_leading\_object\_future\_trajectory}: Provides the predicted future trajectory of the leading object. If no leading object exists, it returns None.
    \item \textbf{get\_future\_trajectories\_for\_specific\_objects}: Returns the future trajectories for specific objects identified by their object IDs. If no such objects exist, it returns None.
    \item \textbf{get\_future\_trajectories\_in\_range}: Retrieves future trajectories where any waypoint falls within a given range (x\_{start}, x\_{end})×(y\_{start}, y\_{end})m². Returns None if no trajectories meet the criteria.
    \item \textbf{get\_future\_waypoint\_of\_specific\_objects\_\\at\_timestep}: Obtains the future waypoints of specific objects at a particular timestep. If an object does not have a waypoint at the specified timestep or if no such object exists, it returns None.
    \item \textbf{get\_all\_future\_trajectories}: Provides the predicted future trajectories for all objects in the scene. Similar to get\_all\_object\_detections, it is advisable to use more specific functions if possible.
\end{itemize}

\paragraph{Occupancy Functions}

\begin{itemize}
    \item \textbf{get\_occupancy\_at\_locations\_for\_timestep}: Determines the probability of occupancy for a list of locations at a specific timestep. Returns None if a location is outside the occupancy prediction scope.
    \item \textbf{check\_occupancy\_for\_planned\_trajectory}: Evaluates whether a planned trajectory collides with other objects.
\end{itemize}

\paragraph{Map Functions}

\begin{itemize}
    \item \textbf{get\_drivable\_at\_locations}: Checks the drivability of specific locations. Returns None if a location is outside the map scope.
    \item \textbf{get\_lane\_category\_at\_locations}: Retrieves the lane category for specific locations. If the location is outside the map scope, it returns None.
    \item \textbf{get\_distance\_to\_shoulder\_at\_locations}: Calculates the distance to both sides of road shoulders for specific locations. Returns None if a location is outside the map scope.
    \item \textbf{get\_current\_shoulder}: Provides the distance to both sides of road shoulders for the current ego-vehicle location.
    \item \textbf{get\_distance\_to\_lane\_divider\_at\_locations}: Computes the distance to both sides of road lane dividers for specific locations. Returns None if a location is outside the map scope.
    \item \textbf{get\_current\_lane\_divider}: Returns the distance to both sides of road lane dividers for the current ego-vehicle location.
    \item \textbf{get\_nearest\_pedestrian\_crossing}: Identifies the location of the nearest pedestrian crossing to the ego-vehicle. Returns None if no pedestrian crossing exists.
\end{itemize}


\section{Implementation Details}\label{sec:app-implement}
In our experiments, we utilized the MS-SWIFT framework to finish our experiment with the following parameter configuration.
\subsection{SFT Phase}\label{sec:sft-exp}
 We trained the model for 20 epochs using the LoRA (Low-Rank Adaptation) fine-tuning method, setting the LoRA rank to 8 and the LoRA alpha to 32. To manage GPU memory usage, the per-device training and evaluation batch sizes were both set to 2, with gradient accumulation enabled over 16 steps. The learning rate was set to $1\times10^{-4}$, and we used the bfloat16 data type to enhance computational efficiency. In order to constrain GPU memory consumption, we froze the parameters of the Vision Transformer (ViT) and set the maximum sequence length to 4096. During training, model evaluation and saving were performed every 1000 steps, with a limit of 5 saved models. Logging was conducted every 5 steps to monitor the training process closely. We employed the DeepSpeed ZeRO-2 optimizer to optimize training performance and disabled the reentrancy of gradient checkpointing for efficiency. Additionally, we configured the warm-up ratio to 0.05 and allocated 4 worker processes for both the data loader and dataset processing.
\subsection{GRPO Phase}\label{sec:grpo-exp}
The RLHF type was designated as "grpo" to utilize the GRPO algorithm for reinforcement learning with human feedback. We activated LoRA training mode(same as SFT Phase in Sec.~\ref{sec:sft-exp}) and set the PyTorch data type to "bfloat16" for efficient computation. The maximum sequence length was configured to 2048, with a maximum completion length of 1024. Training was conducted for 8 epoch, with per-device training and evaluation batch sizes both set to 1. The learning rate was set at $1\times10^{-5}$, and we employed 8 gradient accumulation steps. Model evaluation occurred every 500 steps, with model saving every 100 steps, and we restricted the total number of saved models to 20. Logging was performed at every step to closely monitor the training process. The warm-up ratio for the learning rate was configured to 0.01. We allocated 4 workers to the DataLoader to expedite data loading. For generation, we allowed 2 generations with a temperature of 1.2 to enhance output diversity. The system was prompted with the instruction: "You are a helpful assistant. You first think about the reasoning process in the mind and then provide the user with the answer." We leveraged DeepSpeed's ZeRO-3 stage optimizer to optimize training performance and enabled logging of model completions for further analysis. The beta value was set to 0.001 for specific algorithmic adjustments, and the entire process was conducted for 1 iteration.

\section{Evaluation Metric in the DriveLMM-o1 Benchmark}
\label{sec:Evaluation Metric}

To comprehensively evaluate both quantitative performance and qualitative decision-making capabilities in autonomous driving scenarios, we adopt the evaluation metrics proposed by \cite{ishaq2025drivelmm}.

\textbf{Risk Assessment} examines the model's capacity to prioritize high-risk objects or scenarios, ensuring critical situations receive appropriate urgency in decision-making processes. Accuracy quantifies the precision of environmental perception through correct identification and classification of relevant elements. 

\textbf{Traffic Rule Adherence} measures compliance with established traffic regulations and domain-specific best practices, reflecting real-world operational fidelity. 

\textbf{Scene Awareness} and \textbf{Object Understanding} jointly assess contextual interpretation depth, encompassing accurate perception of spatial relationships, dynamic object behaviors, and complex environmental interactions. 

\textbf{Relevance} evaluates alignment between model outputs and scenario-specific requirements relative to ground truth annotations, ensuring contextually appropriate responses. 

\textbf{Missing Details} identifies critical information gaps through systematic analysis of perceptual blind spots in situational understanding. These complementary metrics collectively establish a holistic framework for evaluating system reliability, safety margins, and environmental adaptability across diverse driving conditions.

\begin{table}[h]
\centering
\small
\begin{tabular}{p{0.27\linewidth} p{0.54\linewidth}}
\toprule
\textbf{Metric} & \textbf{Description} \\
\midrule
Risk Assessment Accuracy & Evaluates if the model correctly prioritizes high-risk objects or scenarios. \\ \midrule
Traffic Rule Adherence & Scores how well the response follows traffic laws and driving best practices. \\ \midrule
Scene Awareness and Object Understanding & Measures how well the response interprets objects, their positions, and actions. \\ \midrule
Relevance & Measures how well the response is specific to the given scenario and ground truth. \\ \midrule
Missing Details & Evaluates the extent to which critical information is missing from the response. \\
\bottomrule
\end{tabular}
\caption{Evaluation metrics for quantitative performance and qualitative decision-making capabilities}
\label{tab:drivelmmo1_metrics}
\end{table}
DriveLMM-o1\cite{ishaq2025drivelmm} utilize GPT-4o-mini to complete the evaluation steps. The prompt we employ is as follows:

\textit{
You are an autonomous driving reasoning evaluator. Your task is to assess the alignment, coherence, and quality of reasoning steps in text responses for safety-critical driving scenarios.
}

You will evaluate the model-generated reasoning using the following metrics:

\begin{enumerate}
    \item \textbf{Faithfulness-Step (1-10)}: Measures how well the model's reasoning steps align with the ground truth.
    \begin{itemize}
        \item 9-10: All steps correctly match or closely reflect the reference.
        \item 7-8: Most steps align, with minor deviations.
        \item 5-6: Some steps align, but several are incorrect or missing.
        \item 3-4: Few steps align; most are inaccurate or missing.
        \item 1-2: Majority of steps are incorrect.
    \end{itemize}
    
    \item \textbf{Informativeness-Step (1-10)}: Measures completeness of reasoning.
    \begin{itemize}
        \item 9-10: Captures almost all critical information.
        \item 7-8: Covers most key points, with minor omissions.
        \item 5-6: Missing significant details.
        \item 3-4: Only partial reasoning present.
        \item 1-2: Poor extraction of relevant reasoning.
    \end{itemize}
    
    \item \textbf{Risk Assessment Accuracy (1-10)}: Evaluates if the model correctly prioritizes high-risk objects or scenarios.
    \begin{itemize}
        \item 9-10: Correctly identifies and prioritizes key dangers.
        \item 7-8: Mostly accurate, with minor misprioritizations.
        \item 5-6: Some important risks are overlooked.
        \item 3-4: Significant misjudgments in risk prioritization.
        \item 1-2: Misidentifies key risks or misses them entirely.
    \end{itemize}
    
    \item \textbf{Traffic Rule Adherence (1-10)}: Evaluates whether the response follows traffic laws and driving best practices.
    \begin{itemize}
        \item 9-10: Fully compliant with legal and safe driving practices.
        \item 7-8: Minor deviations, but mostly correct.
        \item 5-6: Some inaccuracies in legal/safe driving recommendations.
        \item 3-4: Several rule violations or unsafe suggestions.
        \item 1-2: Promotes highly unsafe driving behavior.
    \end{itemize}
    
    \item \textbf{Scene Awareness \& Object Understanding (1-10)}: Measures how well the response interprets objects, their positions, and actions.
    \begin{itemize}
        \item 9-10: Clearly understands all relevant objects and their relationships.
        \item 7-8: Minor misinterpretations but mostly correct.
        \item 5-6: Some key objects misunderstood or ignored.
        \item 3-4: Many errors in object recognition and reasoning.
        \item 1-2: Misidentifies or ignores key objects.
    \end{itemize}
    
    \item \textbf{Repetition-Token (1-10)}: Identifies unnecessary repetition in reasoning.
    \begin{itemize}
        \item 9-10: No redundancy, very concise.
        \item 7-8: Minor repetition but still clear.
        \item 5-6: Noticeable redundancy.
        \item 3-4: Frequent repetition that disrupts reasoning.
        \item 1-2: Excessive redundancy, making reasoning unclear.
    \end{itemize}
    
    \item \textbf{Hallucination (1-10)}: Detects irrelevant or invented reasoning steps not aligned with ground truth.
    \begin{itemize}
        \item 9-10: No hallucinations, all reasoning is grounded.
        \item 7-8: One or two minor hallucinations.
        \item 5-6: Some fabricated details.
        \item 3-4: Frequent hallucinations.
        \item 1-2: Majority of reasoning is hallucinated.
    \end{itemize}
    
    \item \textbf{Semantic Coverage-Step (1-10)}: Checks if the response fully covers the critical reasoning elements.
    \begin{itemize}
        \item 9-10: Nearly complete semantic coverage.
        \item 7-8: Good coverage, some minor omissions.
        \item 5-6: Partial coverage with key gaps.
        \item 3-4: Major gaps in reasoning.
        \item 1-2: Very poor semantic coverage.
    \end{itemize}
    
    \item \textbf{Commonsense Reasoning (1-10)}: Assesses the use of intuitive driving logic in reasoning.
    \begin{itemize}
        \item 9-10: Displays strong commonsense understanding.
        \item 7-8: Mostly correct, with minor gaps.
        \item 5-6: Some commonsense errors.
        \item 3-4: Frequent commonsense mistakes.
        \item 1-2: Lacks basic driving commonsense.
    \end{itemize}
    
    \item \textbf{Missing Step (1-10)}: Evaluates if any necessary reasoning steps are missing.
    \begin{itemize}
        \item 9-10: No critical steps missing.
        \item 7-8: Minor missing steps, but answer is mostly intact.
        \item 5-6: Some important steps missing.
        \item 3-4: Many critical reasoning gaps.
        \item 1-2: Response is highly incomplete.
    \end{itemize}
    
    \item \textbf{Relevance (1-10)}: Measures how well the response is specific to the given scenario and ground truth.
    \begin{itemize}
        \item 9-10: Highly specific and directly relevant to the driving scenario. Captures critical elements precisely, with no unnecessary generalization.
        \item 7-8: Mostly relevant, but some minor parts may be overly generic or slightly off-focus.
        \item 5-6: Somewhat relevant but lacks precision; response contains vague or general reasoning without clear scenario-based details.
        \item 3-4: Mostly generic or off-topic reasoning, with significant irrelevant content.
        \item 1-2: Largely irrelevant, missing key aspects of the scenario and failing to align with the ground truth.
    \end{itemize}
    
    \item \textbf{Missing Details (1-10)}: Evaluates the extent to which critical information is missing from the response, impacting the reasoning quality.
    \begin{itemize}
        \item 9-10: No significant details are missing; response is comprehensive and complete.
        \item 7-8: Covers most important details, with minor omissions that do not severely impact reasoning.
        \item 5-6: Some essential details are missing, affecting the completeness of reasoning.
        \item 3-4: Many critical reasoning steps or contextual details are absent, making the response incomplete.
        \item 1-2: Response is highly lacking in necessary details, leaving major gaps in understanding.
    \end{itemize}
\end{enumerate}

\section*{Final Evaluation}

Compute the Overall Score as the average of all metric scores.

\begin{verbatim}
{
  "Faithfulness-Step": 6.0,
  "Informativeness-Step": 6.5,
  "Risk Assessment Accuracy": 7.0,
  "Traffic Rule Adherence": 7.5,
  "Object Understanding": 8.0,
  "Repetition-Token": 7.0,
  "Hallucination": 8.5,
  "Semantic Coverage-Step": 7.5,
  "Commonsense Reasoning": 7.0,
  "Missing Step": 8.5,
  "Relevance": 8.5,
  "Missing Details": 7.0,
  "Overall Score": 7.42
}
\end{verbatim}

\section*{Guidelines}

\begin{itemize}
    \item Avoid subjective interpretation and adhere to the given thresholds.
    \item Always strictly follow these scoring guidelines.
    \item Do not add any additional explanations beyond the structured JSON output.
\end{itemize}

\begin{figure}[t]
  \centering
  
  \includegraphics[width=\linewidth]{ 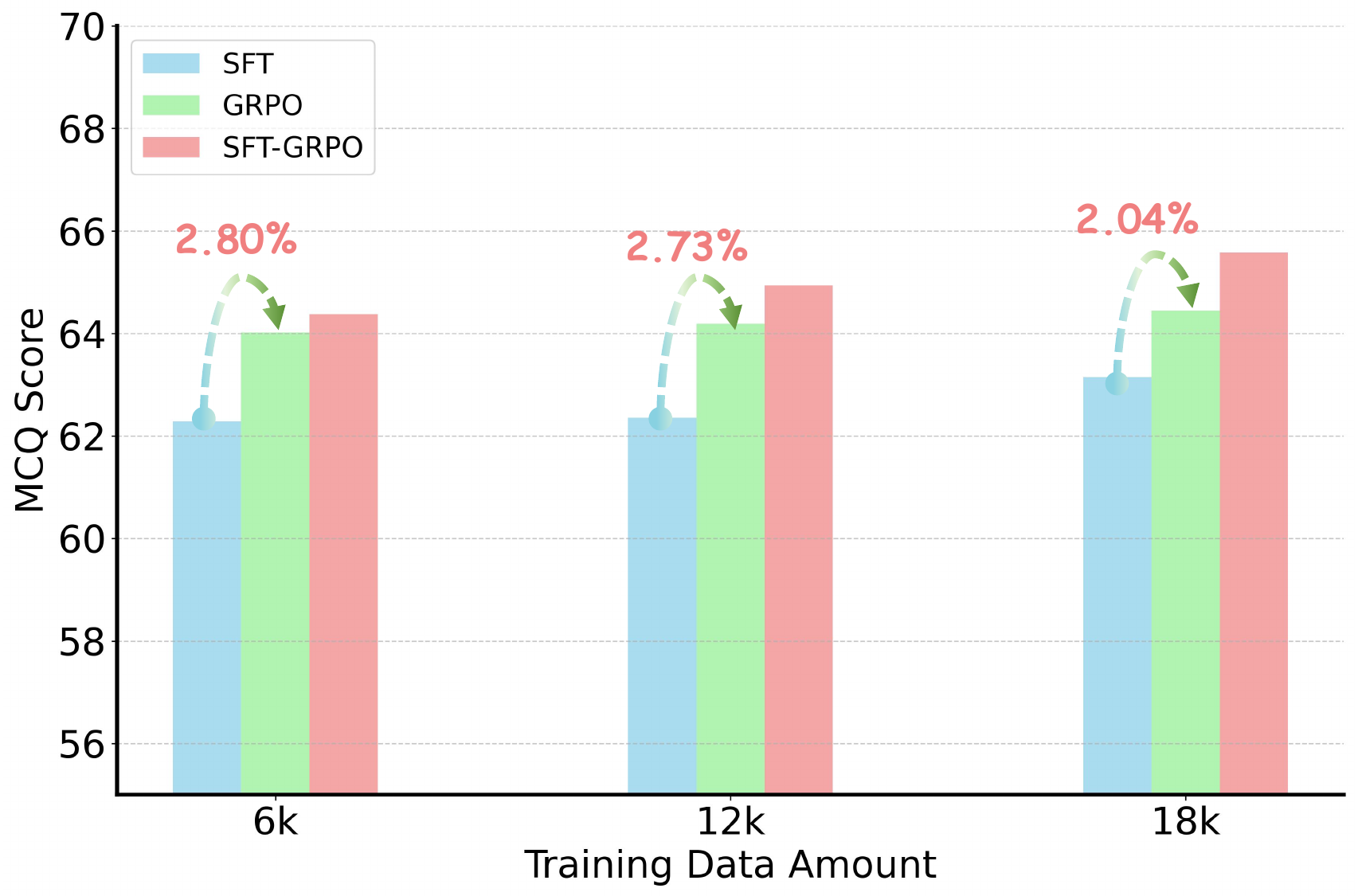}
  \caption{Our GRPO method with structured rewards setting and two-stage training strategy significantly boost final answer accuracy. With just 6k samples, it outperforms SFT, showing strong performance even with limited data. As data increase, AgentThink consistently leads in MCQ performance.}
  \label{fig:data-ablation}
\end{figure}

\begin{table*}[!ht]

\footnotesize 
\resizebox{\linewidth}{!}{
\begin{tabular}{l|l|ccc|cc|cc}
    \toprule
    \multirow{2}{*}{ID} & \multirow{2}{*}{Amount of Data} 
        & \multicolumn{3}{c|}{Driving Metrics (\%) $\uparrow$} 
        & \multicolumn{2}{c|}{Scene Detail (\%) $\uparrow$} 
        & \multicolumn{2}{c}{Overall (\%) $\uparrow$} \\ 
        & & Risk Ass. &  Rule Adh. &Obj. Und. 
          & Relevance & Missing 
          & Reason. & MCQ \\ 
    \midrule
    (a) & GRPO w/ 6k data  & 68.79 & 74.93 & 71.12 & 75.44 & 67.80 & 69.09 & 64.02 \\
    (b) & GRPO w/ 12k data & \textbf{69.26} & \textbf{75.44} & \textbf{71.68} & \textbf{75.85} & \textbf{68.13} & \textbf{69.48} & \textbf{64.19} \\
    \midrule
    (c) & SFT w/ 6k data   & 74.47 & 80.58 & 75.82 & 80.66 & 72.02 & 74.08 & 62.28 \\ 
    (d) & SFT w/ 12k data  & \textbf{74.81} & \textbf{80.88} & \textbf{76.21} & \textbf{80.90} & \textbf{72.42} & \textbf{74.33} & \textbf{62.36} \\
    \midrule
    (e)	 & SFT-GRPO w/ 6kdata	 &75.60  &	81.40 	 &76.97  &	81.74 	 &72.51 	 &74.92  &	64.38\\
     (f)	& SFT-GRPO w/ 12kdata	& \textbf{75.69} &	\textbf{82.29}  &	\textbf{76.98}  &	\textbf{82.58}  &	\textbf{72.39}  &	\textbf{75.08} 	 & \textbf{64.94} \\
    \bottomrule
\end{tabular}}
\vspace{-10pt}
\caption{Impact of training data amount on key evaluation metrics (reordered by driving metrics first).}
\vspace{-12pt}
\label{tab:ablation_v2}
\end{table*}

\section{Visualization of Main Experiment }
\label{sec:Visualization}
\begin{figure*}[t]
  \centering
  \includegraphics[width=0.8\linewidth]{ 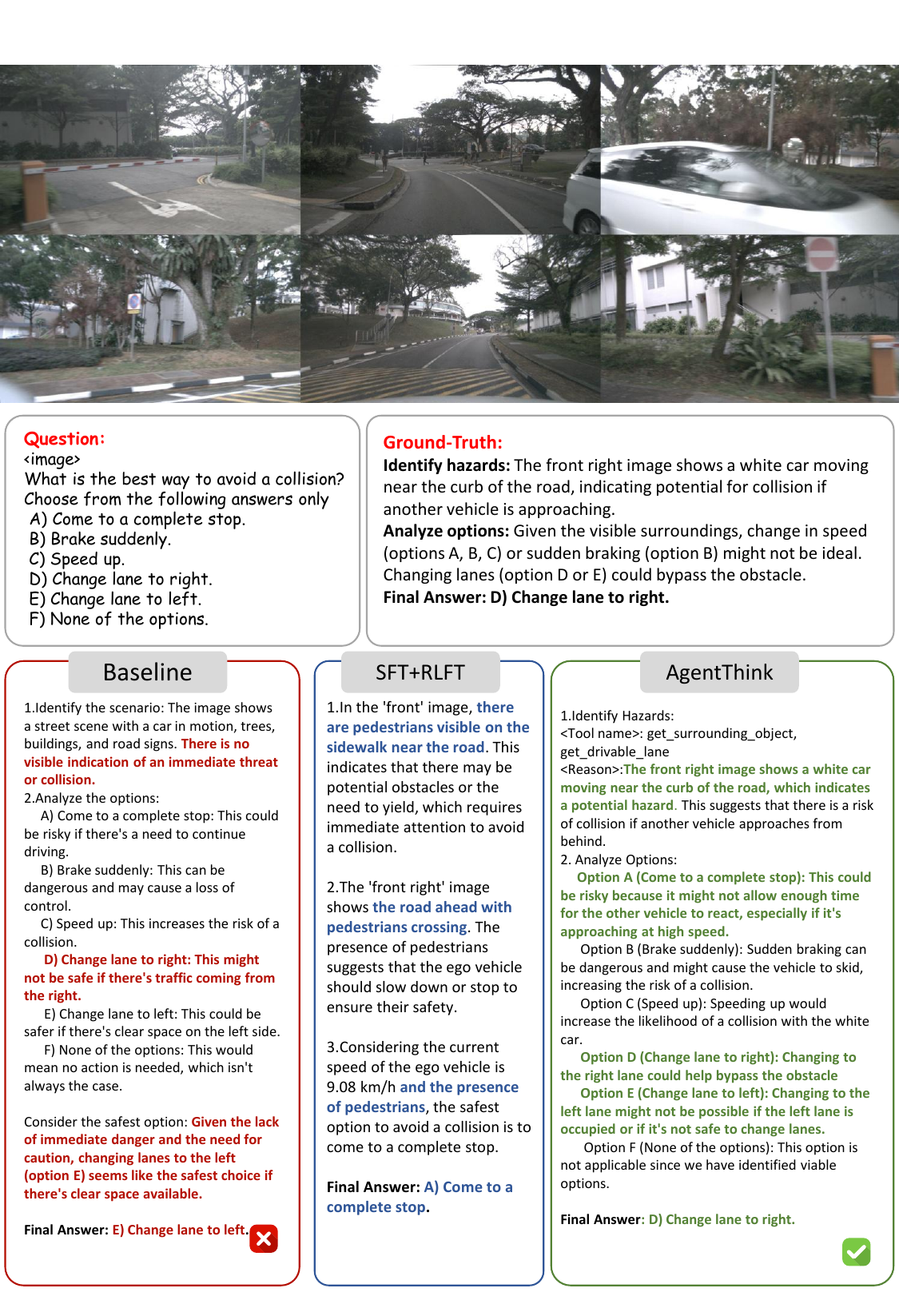}
  \caption{Visulization of Results. Correct scene understanding outputs are highlighted in green, while erroneous interpretations are marked in red for comparative analysis. Notably, AgentThink demonstrates accurate analytical capabilities and avoid overly conservative decision-making tendencies (e.g. unnecessary braking), which compromise driving comfort despite maintaining safety margins. }
  \label{fig:vis_1}
\end{figure*}

\begin{figure*}[t]
  \centering
  \includegraphics[width=0.8\linewidth]{ 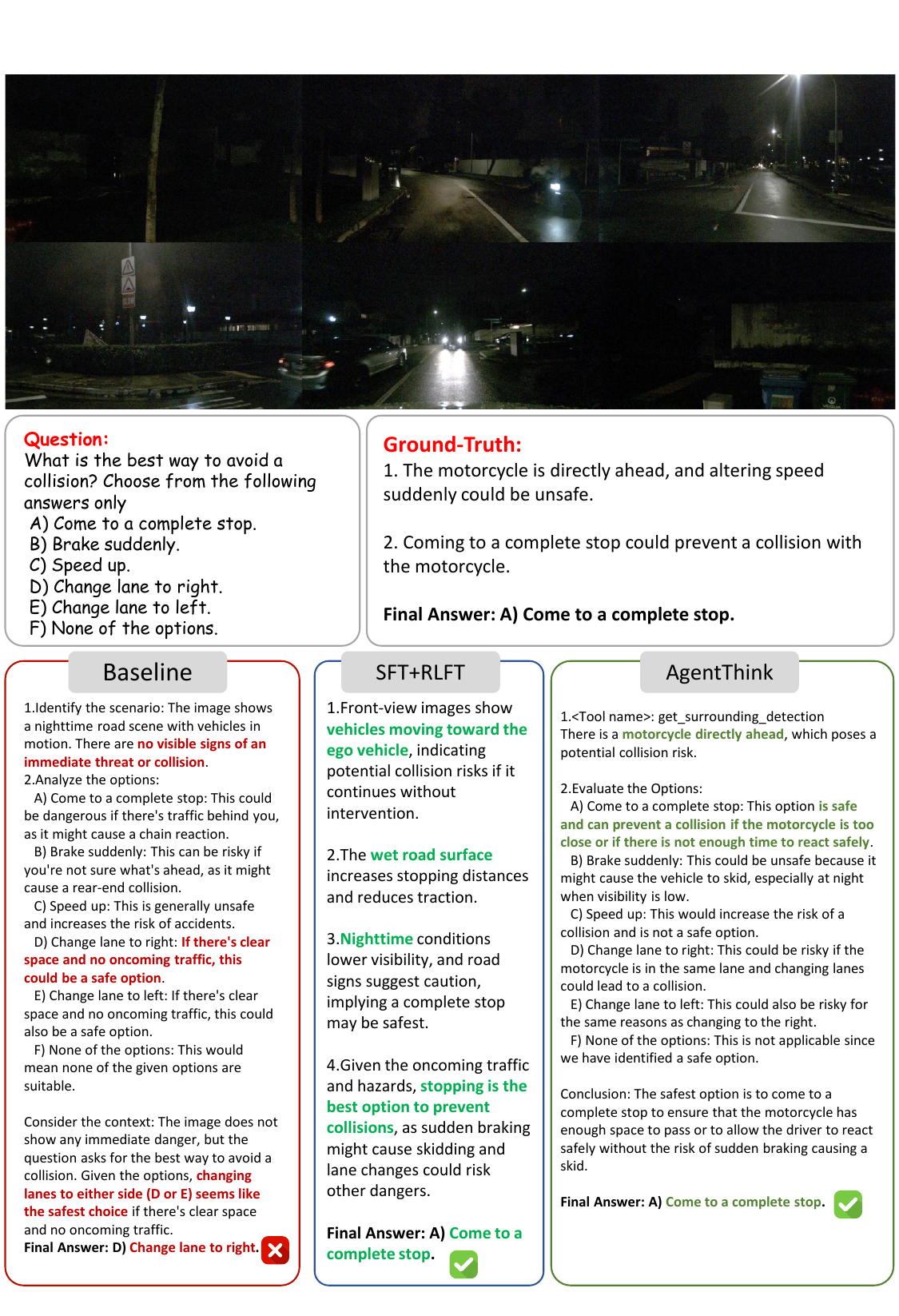}
  \caption{Visulization of AgentThink. Correct scene understanding outputs are highlighted in green, while erroneous interpretations are marked in red for comparative analysis.}
  \label{fig:vis_2}
\end{figure*}

Here we provide the visualizations for different scenarios (Fig \ref{fig:vis_1}, \ref{fig:vis_2}). Correct scene understanding outputs are highlighted in green, while erroneous interpretations are marked in red for comparative analysis. Notably, AgentThink demonstrates accurate analytical capabilities and avoid overly conservative decision-making tendencies (e.g. unnecessary braking), which compromise driving comfort despite maintaining safety margins.

\section{Impact of Data Size}
\label{sec: Impact}
To systematically evaluate the influence of training data scale, we conduct ablation studies on Qwen2.5-VL-7B with fixed vision encoder parameters. The base model is first finetuned via LoRA-based SFT, followed by policy optimization through GRPO algorithm. All experiments are implemented on 16× NVIDIA A800 GPUs with unified hyperparameter configurations. As presented in Tab.~\ref{tab:ablation_v2} and Fig.~\ref{fig:data-ablation}, reducing training data size from 12k to 6k samples leads to a performance decline across all metrics, with SFT-based training exhibiting a more pronounced degradation. Notably, RLFT demonstrates superior robustness to data scarcity, achieving statistically significant performance gains over SFT counterparts in low-data regimes ($52.32\%$ vs. $62.28\%$ final accuracy at 6k data scale). This suggests the inherent advantage of reinforcement learning frameworks in leveraging limited training samples for vision-language policy learning.

\section{Evaluation Metric in the DriveMLLM Benchmark}
\label{sec: Evaluation DriveMLLM}
The eight metrics in the DriveMLLM benchmark \cite{guo2024drivemllm} can be described as follows:

\paragraph{Left/Right (L/R)}
This metric evaluates the model's ability to identify which of two objects is positioned further to the left or right in the image based on their inferred \(x\)-coordinates.

\paragraph{Front/Back (F/B)}
This assesses whether the model can determine if one object is in front of or behind another using depth cues, reflecting its understanding of \(z\)-coordinate relationships.
\begin{equation}
acc_i = 
\begin{cases} 
1, & \text{if } p_i = y_i \\
0, & \text{if } p_i \neq y_i 
\end{cases}
\end{equation}

where:
- \( p_i \) is the model's predicted label for sample \( i \).
- \( y_i \) is the ground truth label for sample \( i \).

The overall accuracy \( acc \) is calculated as the mean of the individual accuracies.
\paragraph{Relative Height Difference (RHD)}
It tests the model's capability to calculate the vertical distance between the camera and an object based on the object's \(z\)-coordinate.

\paragraph{Relative Distance (RD)}
This metric examines the model's ability to estimate the Euclidean distance from the camera to a specified object using inferred spatial information.
\begin{equation}
acc_i = \frac{1}{1 + \alpha_d \|d_i - d_i^{\text{gt}}\|_1}
\end{equation}

where:
- \( d_i \) is the model's predicted distance for sample \( i \).
- \( d_i^{\text{gt}} \) is the ground truth distance for sample \( i \).
- \( \alpha_d \) is a scaling factor controlling the penalty for deviation, set to \( \alpha_d = 0.05 \).
\paragraph{Positional Precision (PPos)}
It evaluates the model's accuracy in identifying the exact coordinates \([x, y]\) of a specified object within an image.
\begin{equation}
acc_i = \frac{1}{1 + \alpha_p \|\mathbf{x}_i - \mathbf{x}_i^{\text{gt}}\|_2}
\end{equation}

where:
- \(\mathbf{x}_i = (x_i, y_i)\) are the model's predicted coordinates for sample \(i\).
- \(\mathbf{x}_i^{\text{gt}} = (x_i^{\text{gt}}, y_i^{\text{gt}})\) are the ground truth coordinates for sample \(i\).
- \(\alpha_p\) is a scaling factor, set to \(\alpha_p = 0.005\).

\paragraph{Bounding Box (BBox)}
This metric assesses the model's capability to determine the bounding box coordinates \([min\_x, min\_y, max\_x, max\_y]\) of a specific object.
\begin{equation}
acc_i = \text{IoU}(B_i, B_i^{\text{gt}}) = \frac{|B_i \cap B_i^{\text{gt}}|}{|B_i \cup B_i^{\text{gt}}|}
\end{equation}

where:
- \( B_i \) is the model's predicted bounding box for sample \( i \).
- \( B_i^{\text{gt}} \) is the ground truth bounding box for sample \( i \).
- \( |\cdot| \) denotes the area of the bounding box.

\paragraph{Camera-Object Vertical Distance (CVD)}
It tests the model's ability to calculate the vertical distance from the camera to an object based on the object's \(z\)-coordinate.

\paragraph{Camera-Object Euclidean Distance (CD)}
This metric evaluates the model's skill in computing the Euclidean distance between the camera and an object using inferred spatial information.
The computation of the CVD and CD metrics follows the same methodology as that of the RVD and RD metrics.
\paragraph{Overall Accuracy Score (AccS)}
This provides a comprehensive measure of the model's performance across all the aforementioned tasks, reflecting its overall spatial understanding capabilities.
\begin{equation}
    acc = \frac{1}{N} \sum_{i=1}^{N} acc_i \quad \text{AccS} = \frac{1}{8} \sum_{j=1}^{8} acc_j
\end{equation}

\end{document}